\documentclass[%
 reprint,
 amsmath,amssymb,
 aps,
]{revtex4-2}

\usepackage{graphicx}% Include figure files
\usepackage{dcolumn}% Align table columns on decimal point
\usepackage{bm}% bold math
\begin{document}

\preprint{APS/123-QED}

\title{Mechanics-based Analysis on Flagellated Robots}% Force line breaks with \\

\author{Yayun Du$^{1}$, Andrew Miller$^{1}$}
%  \altaffiliation{Physics Department, XYZ University.}%Lines break automatically or can be forced with \\
\author{M. Khalid Jawed$^{1,}$}%
 \email{Correspondence should be addressed to khalidjm@seas.ucla.edu}
\affiliation{%
 $^{1}$Department of Mechanical \& Aerospace Engineering, \\ University of California, Los Angeles
}%

% \collaboration{MUSO Collaboration}%\noaffiliation

% \author{Yayun Du}
% %  \homepage{http://www.Second.institution.edu/~Charlie.Author}
% \affiliation{
%  University of California, Los Angeles% with \\
% }%
% \affiliation{
%  Third institution, the second for Charlie Author
% \affiliation{%
%  Authors' institution and/or address\\
%  This line break forced with \textbackslash\textbackslash
% }%

% \collaboration{CLEO Collaboration}%\noaffiliation


\begin{abstract}
We explore the locomotion of soft robots in granular medium resulting from the elastic deformation of slender rods. A low-cost, rapidly fabricable robot is presented that is inspired by the physiological structure of bacteria. It consists of a rigid head, with a motor and batteries embedded, and multiple elastic rods -- our model for flagella -- to investigate locomotion in granular media. The elastic flagella are rotated at one end by the motor, and they deform due to drag from the granular medium, causing the robot to propel forward. The external drag is determined by the flagellar shape, while the flagellar shape changes due to the competition between external loading and elastic forces. In this coupled fluid-structure interaction problem, interestingly, we observe that -- depending on the physical parameters of the system -- increasing the number of flagella can decrease (design 1) or increase (design 2) the propulsive speed of the robot. This nonlinearity in the functional relation between propulsion and the physical parameters of this simple robot motivates us to fundamentally analyze its mechanics using theory, numerical simulation, and experiments. We present a simple Euler-Bernoulli beam theory-based analytical framework that is capable of qualitatively capturing both designs. Theoretical prediction quantitatively matches experimental data when the deformation in the flagella is small. To account for the geometrically nonlinear deformation that is often encountered in soft robots and microbes, we implement a simulation framework that incorporates discrete differential geometry-based simulations of elastic rods, a resistive force theory-based model for drag, and a modified Stoke\rq{}s law for the hydrodynamics of the robot head. Comparison with experimental data indicates that the simulations can quantitatively predict the robotic motion in both designs. Overall, the theoretical and numerical tools presented in this paper can shed light on the design and control of this class of articulated robots in granular or fluid media.
\begin{description}
\item[Keywords]
flagellar locomotion, granular medium, Euler-Bernoulli beam theory, fluid-structure interaction, design and modeling, soft robotics
\end{description}
\end{abstract}
% \keywords{flagellar locomotion, granular medium, Euler-Bernoulli beam theory, fluid-structure interaction, design and modeling, soft robotics }%Use showkeys class option if keyword
\maketitle

%\tableofcontents

\section{\label{sec:intro}Introduction}
Apart from animals capable of crawling, digging~\cite{hosoi2015beneath}, slithering~\cite{hu2009mechanics}, swimming~\cite{lauder2007fish}, and gliding~\cite{dickinson2000animals, ma2013controlled} in environments, bacteria, one major group of microorganisms, also inspire the development of novel and efficient robots. 90\% of marine bacteria\cite{leifson1964motile} achieve efficient locomotion in a fluid through the rotation of a flagellum (a slender filament), exploiting the anisotropic drag to produce thrust and violating the constraints of the scallop theorem\cite{lauga2011life}. 
% The rotation of a single or multiple flagella in a fluid medium is the primary mode of locomotion for bacteria. 
Flagellar locomotion results from a non-trivial coupling between the geometrically nonlinear deformation in the flagellum and mechanics of the surrounding medium, posing a challenging fluid-structure interaction (FSI) problem.
% FSI leads to a variety of flagellar propulsion such as tumbling, turning, polymorphic transformations and bundling formation. 

Theoretical study about flagellar propulsion dates back to 1955 when Taylor \cite{taylor1951analysis} first analyzed the swimming of microscopic organisms. Over the last two decades, numerous studies have studied flagellar propulsion in low Reynolds fluids through experiments\cite{turner2000real, kudo2005asymmetric, fujii2008polar, son2013bacteria, thawani2018trajectory, noselli2019swimming}, computation\cite{ cortez2001method,kim2005deformation, vogel2012motor}, and theory\cite{ berke2008hydrodynamic,spagnolie2012hydrodynamics, lopez2014dynamics, armanini2021flagellate}. Recent investigations\cite{du2021modeling, du2021simple} have modelled the flagellum as a Kirchhoff elastic rod\cite{kirchhoff2020gleichgewicht}, and coupled to the fluid with hydrodynamic forces\cite{lighthill1976flagellar}. Jawed investigated the dynamics of a helical elastic flagellum rotating in a viscous fluid\cite{jawed2015propulsion} and near a rigid boundary\cite{jawed2017dynamics}. However, the role of the head and the flow generated by its motion and coupling with the flagellum-induced flow are ignored. Huang performed simulations that utilized flagellar buckling to change moving direction on a robot composed of a mass-point head and uni-flagellum\cite{huang2020numerical}.  Our previous work\cite{du2021simple} established an untethered articulated robot that was composed of a rigid head and multiple soft flagella and used discrete differential geometry to simulate the flagella and resistive force theory\cite{lighthill1976flagellar} (RFT) to model the interaction force between the fluid and flagella. Simulation and experimental results agreed well quantitatively. Our robot also demonstrated the same behavior as bacteria, ii.e., head and flagella rotating in the opposite directions\cite{lauga2006swimming} and circling when near the air-fluid boundary.  

\begin{figure*}[t!]
    \centering
    \includegraphics[width=\textwidth]{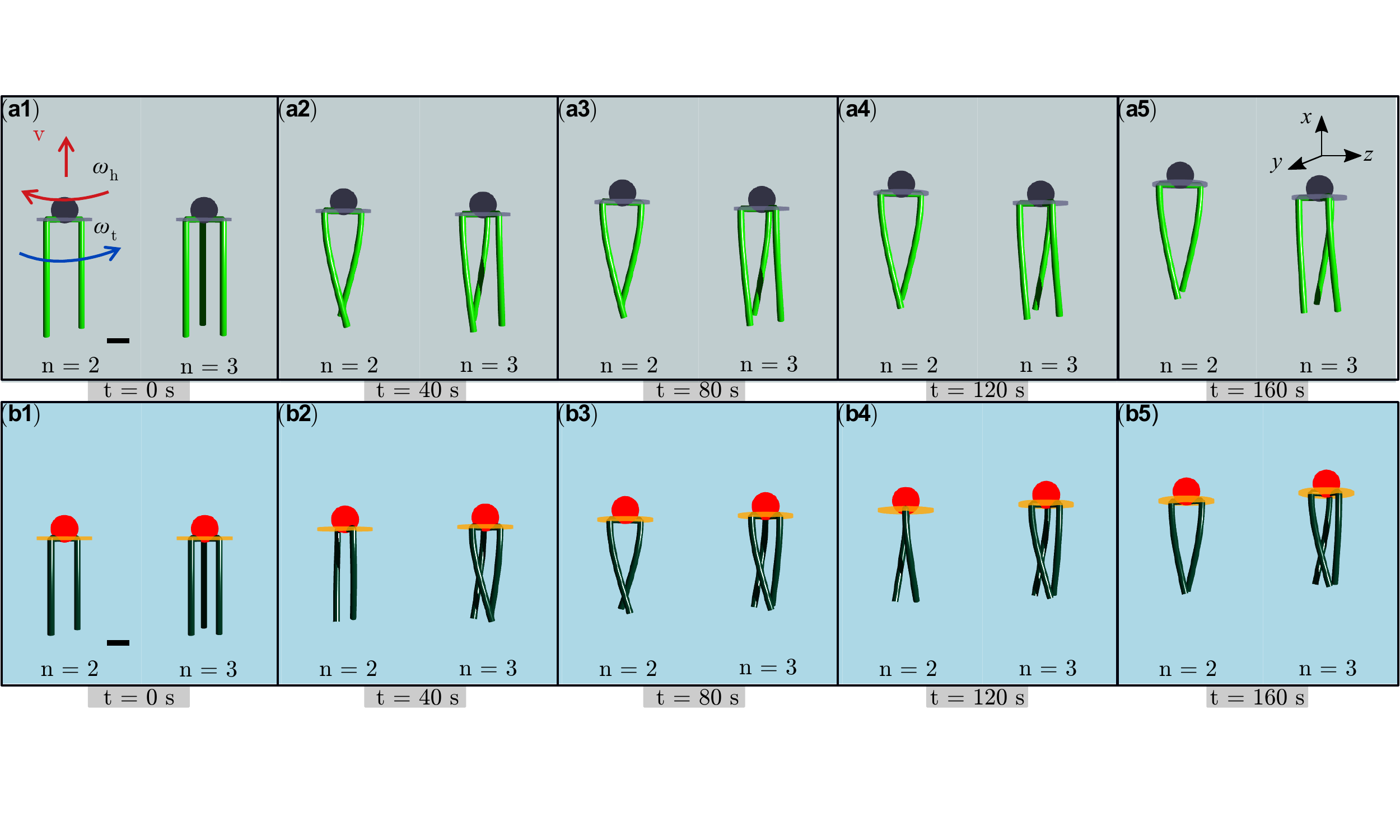}
    \caption{Snapshots from simulation. The first row depicts the shape and movement of the robot in design 1 where the robot speed decreases as the number of flagella increases from, $n = 2$, to $n = 3$ in panels (\textbf{a1}-\textbf{a5}). In contrast, the second row illustrates the shape and movement of the robot in design 2 where as the number of flagella, $n$, increases from two to three in panels (\textbf{b1}-\textbf{b5}), the speed of the robot increases. In (\textbf{a1}-\textbf{a5}), the total rotational speed of the motor $\omega_T=100.00$ rpm, head rotational speed  $\omega_h=95.47$ rpm, and flagellar rotational speed $\omega_t=4.53$ rpm while for $n = 3$ in (\textbf{a1}-\textbf{a5}), $\omega_T=100.00$ rpm, $\omega_h=97.49$ rpm, and $\omega_t=2.51$ rpm. In (\textbf{b1}-\textbf{b5}) and $n=2$, $\omega_T=100.00$ rpm, $\omega_h=44.06$ rpm, and $\omega_t=55.94$ rpm while  $\omega_T=100.00$ rpm, $\omega_h=56.98$ rpm, and $\omega_t=43.02$ rpm for $n = 3$ in (\textbf{b1}-\textbf{b5}). Black bar is 2 cm.
    Physical parameters are available in Sec.\ref{subsec:parameters}.
    }
    \label{fig:simulationSnapshots}
\end{figure*}

RFT, initially used for analyzing the movements of microscopic organisms in viscous fluids, is  proved to apply to animal and robot locomotion on and within granular media (GM) \cite{maladen2011undulatory}. This builds an intimate link between the microscopic bacterial world and meter-sized animals in sand. The most recent studies \cite{texier2017helical, agarwal2021surprising},
on rotational intruders moving through GM conclude on the empirical feasibility of applying modified and entirely empirical RFT, e.g. granular RFT \cite{agarwal2021surprising}. These experimental studies demonstrate that RFT is simple yet very effective in granular materials considering the complicated constitutive features of granular matter such as nonlinearity and nonlocality. However, the intruders in these works are \emph{rigid}, inspiring our further study on the geometrical design and efficiency study on robots with \emph{soft} flagella in GM. Previously, our other work modeled the locomotion of articulated soft robots in GM\cite{du2021modeling}. Leading design and control parameters of our untether robots are the rotational speed of the embedded motor, the number of flagella, geometrical parameters of flagella such as radius and length. Some counter-intuitive observations include the inverse relationship between the speed and the number of flagella of the robot in the representative setup (as shown in Fig.\ref{fig:simulationSnapshots}(\textbf{a1-a5})). This verifies that the flagellar locomotion is the result of a complicated coupling between granular mechanics and deformable bodies, a complex FSI problem.

In summary, while there exist works\cite{thawani2018trajectory, huang2020numerical, du2021simple} that explore flagellar locomotion considering the effect of the head, systematic simulations are required for more quantitative predictions, e.g., under what design and control parameters the maximal efficiency of actuation is achieved. Furthermore, although a comprehensive simulation framework is offered in our earlier work\cite{du2021modeling}, elaborate equations suffocate and make it difficult for designers to understand and predict the movement of the system directly. As a result, a reduced model capable of qualitatively capturing the relationship between the performance of flagellar locomotion in GM versus the design space is required, hence avoiding the cumbersome trial and error design process.

Here, we employ a structural robot design similar to that used in our previous work\cite{du2021modeling}: a palm-sized untethered robot composed of  $n\geq 2$ naturally straight elastic rods and a rigid head housing a geared dc motor and batteries. As illustrated in Fig.~\ref{fig:simulationSnapshots}, the rotation of flagella induces the drag force from the surrounding GM because of their flexibility. Flagellar deformation provides a net propulsive force forward, which is zero in the case of straight \emph{rigid} flagella. 
 Meanwhile, we implement two numerical frameworks, both of which employ Resistive Force Theory (RFT) to model the drag on a collection of multiple elastic rods and modified Stoke\rq{}s law to calculate the drag force and moment on the head. In contrast, one uses Euler-Bernoulli beam theory, and the other applies discrete differential geometry (DDG) to simulate the flagellated robot structure. When the motor rotates slowly, e.g., $\omega_{\textrm{T}}$ $\leq$ 10 rpm, the deformation of the robot flagella is linear, and the resulting propulsive force calculated using DDG simulation is nearly identical to that obtained from Euler-Bernoulli beam theory. However, the beam theory fails to precisely capture the nonlinear deformation of the robot flagella at high motor speeds, e.g., $\omega_T = 250$ rpm, DDG-based framework is still capable of accurately representing the performance of the robot in experiments. Due to the rod-based kinematic representation of the robot, this computational tool is used to simulate flagellated robots even faster than real-time. Additionally, our prior study uncovered a counterintuitive phenomenon, the speed of the robot falls as the number of flagella rises (design 1), which is captured by both the beam theory and DDG-based frameworks. However, the beam theory-based framework predicts the existence of design 2, in which the speed of the robot increases as the number of flagella grows. This prediction is successfully confirmed experimentally after we modify the robot design and radius of the GM.
In summary, the simple overall Euler-Bernoulli beam theory-based framework can qualitatively capture both designs of the complicated locomotion in GM, the intricate interaction between the robot and GM. In contrast, the DDG-based framework is capable of quantitatively simulating the complex locomotion in GM.

\section{\label{sec:problemDefinition}Problem Statement}
In this work, we analyze flagellated locomotion in GM, a typical FSI problem, using a mechanics-based approach. As a result, both the experimental and theoretical aspects of the problem involve two primary components, the GM (fluid part) and the robot (structure part). A GM must be selected to complement the architecture of the untethered flagellated robot. Internal friction between granules prevents the robot head from fluidizing the granules in front and propelling forward, whereas insufficient friction results in ineffective flagellar propulsion. Internal friction during locomotion is correlated with the phase transition of GM from solid to fluid, which is governed by temperature and volume fraction (VF)\cite{gravish2014effect} - the ratio of solid to occupied volume. The less effort required to modify the VF of GM, the better for investigating the locomotion in GM. Apart from the volume fraction of the GM, other physically controllable parameters include the shape and material friction of the head, as well as the number, radius, length, and stiffness of the flagella and the rotational speed of the motor. Among them, the number, $n$, and the rotational speed of the motor, $\omega_T$, are the most precisely trackable with the least effort owing to the design simplicity of the robot and experiments. Their effect on the translational speed of the robot, $v$, is explored and will be displayed in Sections~\ref{subsec:speedOfTheRobot} and ~\ref{subsubsec:VelocityVSFlagellaNumber}. 

The entire robot structure consists mainly of active multiple soft elastic flagella and a rigid passively actuated head, all of which rotate along the long axis of the robot. The case where the flagella are fixed at a 3D-printed plate with uniform external drag forces applied is analogous to the case of a cantilever beam with a uniformly distributed load.
In what follows, first, we outline the Euler-Bernoulli beam theory-based mechanics analysis of the untethered robot, including (1) the external loading from the GM onto flexible flagella and (2) the drag force on the head. Beam theory qualitatively captures the two cases in which increasing $n$ can either accelerate or retard the speed, $v$. Whichever happens, in reality, is determined by the intricate balance of the competition between the external loading and elastic forces. According to experimental evidence, the result is closely related to the robot head and motor speed design, $\omega_T$.
Second, we introduce a numerical model of the robot in which the robot structure is represented by a network of Kirchhoff's rods\cite{}. Finally, we present experiments conducted to quantify the propulsive speed of this class of flagellated robots in Section~\ref{sec:experiment}.

\begin{figure}
   \centering
   \includegraphics[width=\columnwidth]{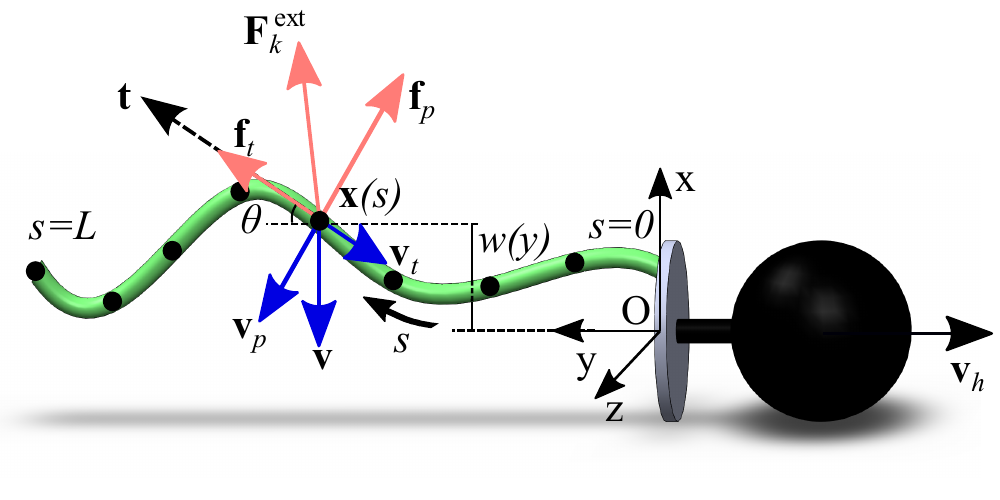}
   \caption{Schematic showing the velocities and induced resistive forces on one flagellum rotating at a constant rotational speed $\omega_ t$ on the robot moving forward at a velocity of $\textbf{v}_ h$. The coordinate system ($O-x-y-z$) is fixed at the center of the 3D-printed plate; the flagellum starts at $s = 0$, propagates along the $x$-direction until the end ($s = L$), and is divided into a series of nodes. Each node (numbered based on differential geometry) is characterized by its tangential direction $\textbf{t}$ and its velocity $\textbf{v}$ direction; $k$-th node (with coordinate  $\textbf{\textrm x}(s)$) experiences a force $\textbf{\textrm F}^{\textrm{ext}}_k$ that propels the robot forward (along $-y$-direction). 
   }
   \label{fig:schematic}
\end{figure}

\section{Methodology}
\label{sec:methodology}

\subsection{Beam theory based analysis of propulsion}
\label{sec:beam}

Benefiting from the elasticity property, the soft flagella of the robot bend out of the long axis, and the drag force generated by the surrounding granules produces a component in the -$y$ direction as displayed in Fig.~\ref{fig:schematic}, propelling the robot forward.

\subsubsection{Clamped flagellum fixed in space}
\label{subsubsec:flagellarPropulsion}
% The deformation of the elastic rod can be modelled using one of two analytical approaches - with the linear assumption, which neglects some small terms and assumes small deformation, or with the nonlinear solution, which does not make any of these assumptions.  
For illustration, one flagellum of the robot in a discrete setting moving in the GM is displayed in Fig.~\ref{fig:schematic}. Hereafter, unless otherwise stated, all the parameters, e.g. forces, are associated with one flagellum. Consider a node $\mathbf x(s)$ along the flagellum, where $s \in [0, L]$ is the arc-length parameter and $L$ is the length of the flagellum. At steady state, the magnitude of the velocity vector at $\mathbf x(s)$ is defined as 
\begin{equation}
 v = \|\mathbf v\| = \omega_{{t}} R_{\textrm{d}},   
 \label{eq:tail_velocity}
\end{equation}
where $\mathbf v$ is along $-y$ axis, $\omega_{{t}}$ stands for the rotational speed of flagella and $R_{\textrm{d}}$ is the radius of 3D printed plate holding soft flagella. The velocity, $\mathbf v$ at 
this point can be decomposed into two parts: the parallel term $ \mathbf{v}_t = (\mathbf v \cdot \mathbf t) \mathbf t = v_t \mathbf t$  and the perpendicular term $ \mathbf{v}_p = \mathbf{v} - \mathbf{v}_t$, where $v_t = -v \sin \theta$ $\mathbf t$ and $\mathbf t$ represents the tangent at the point, whose direction is shown in Fig.~\ref{fig:schematic}. Inspired by Maladen \textit{et al.} \cite{maladen2011undulatory} and Texier \textit{et al.} \cite{texier2017helical}, the drag force exerted by the GM on the robot is modeled using generalized  Coulomb\rq{}s friction law in which the friction force experienced by a slender cylinder is broken down into components normal and tangential to the segment axis, with two corresponding force coefficients. The drag is shown to be dependent on the shape of the object, e.g. the diameter\cite{albert2001granular} . 
Additionally, one work\cite{coq2008rotational} considers the effect of the geometry of the soft filament while modeling the local drag. As a result, the tangential and perpendicular force constants (force per unit length) from the medium that resist $\mathbf{v}_t$ and $\mathbf{v}_p$ are
\begin{subequations}
\begin{align}
\mathbf f_t &= - \eta_{t} \mathbf v_t,\\
\mathbf f_p &= - \eta_{p} \mathbf v_p,
\end{align}
\label{eq:externalForceTail}
\end{subequations}
where the drag coefficients, one along the tangential direction and one along the perpendicular direction are
\begin{subequations}
\begin{align}
\eta_{t} &= \frac{2 \pi \mu}{\log(\frac {2L} {r_0}) - \frac {1} {2}}  \\
\eta_{p} &= \frac{4 \pi \mu}{ \log(\frac {2L} {r_0}) + \frac {1} {2}} , 
\end{align}
\end{subequations}
where $\mu$ is the coefficient constant that quantifies the robot-granule friction as a function of the granule size and the inertia and surface friction of the head, and $L$ and $r_0$ are the length and radius of each flagellum, respectively. 
The drag force constant of external forces at node $\mathbf x(s)$ is \begin{equation}
    \mathbf F^{\textrm{ext}} = \mathbf f_t + \mathbf f_p.
\label{eq:viscousForce}
\end{equation}
Then, the component of $\mathbf F^{\textrm{ext}}$ along the $x$-axis (i.e. antiparallel to velocity) is 
\begin{align}
    p_t
    % & = - \eta_{t} v_{t} \left( \mathbf t  \cdot x \right) - \eta_{p} \left( \left( \mathbf v  \cdot x \right) - v_{t} \left( \mathbf t  \cdot x \right) \right),\\
    % \implies p_t 
    = - \eta_{t} v_{t} \sin \theta - \eta_{p} \left( - v - v_{t} \sin \theta \right),
    \label{eq:forceT}
\end{align}
where 
\begin{equation}
    \sin \theta = \frac{  \mathrm d w}{\sqrt{\mathrm d y^2 + \mathrm dw^2}},
    \label{eq:sin_theta}
\end{equation}
and 
\begin{equation}
    \cos \theta = \frac{  \mathrm d y}{\sqrt{\mathrm d y^2 + \mathrm dw^2}}.
    \label{eq:cos_theta}
\end{equation}
The transverse displacement of a flagellum, which is modeled as a beam is $w (s)$. Assuming a small deflection, $w(s) \approx w(y)$ where $w(y)$ represents the flagellar deflection along the $y$-axis. 
We ignore the effects of head translation in this section; they will be discussed in the following section.
% Hereafter, the subscripts $\left[ \right]_{\textrm{t}}$ and $\left[ \right]_{\textrm{h}}$ denotes the term related to flagella and head, respectively. 

The viscous force can be split into $y$- and $x$-components; the former produces propulsive force, while the latter produces a torque that rotates the head. Given our definition of the direction of $y$-axis, the force constant of propulsive force is as follows along $-y$:
\begin{equation}
q
% = - \mathbf F^{\textrm{ext}} \cdot \hat y 
=  \left( \eta_{p} - \eta_{t} \right) \omega_{{t}} R_{\textrm{d}} \sin \theta \cos \theta.
    \label{eq:longitudinalForce}
\end{equation}

{ % May not need this.
% Next, we simply integrate $q_t$ in Eq.~\ref{eq:longitudinalForce} along the flagellar length to find the propulsive force:
% \begin{equation}
%     F_p = \int_{y=0}^{y=L} q_t(y) \mathrm{d} y
%     \label{eq:Fp}.
% \end{equation}
Without ignoring any higher order terms, i.e. $\sin^2\theta$, the $x$-axis force constant is formulated:

\begin{equation}
p_{t} = EI \frac{\mathrm{d}^4 w}{\mathrm d y^4} = (\eta_{t} - \eta_{p}) v \sin^2 \theta + \eta_{p} v,
\label{eq:beamBending_withoutHead}
\end{equation}
where $EI$ is the bending stiffness of the beam (i.e. the flagellum).
%
% We likewise integrate this across the length of the flagella to get the force in the $x$-direction:

% \begin{equation}
%     F_x = \int_{y=0}^{y=L} p_t(y) \mathrm{d} y
%     \label{eq:Fx}.
% \end{equation}
}

\subsubsection{Clamped flagellum moving at constant speed}
\label{subsubsec:headeffect}
During the robotic movement, in addition to the flagella, its head is also rotating and translating. Assuming the robot moves along the $y$-axis at speed $v_{{h}}$, we rewrite its velocity as 

\begin{equation}
    \mathbf v = - \omega_{{t}} R_{\textrm{d}} \hat x - v_{{h}} \hat y,
    \label{eq:accurateVelocity}
\end{equation}
where $\hat x, \hat y$ are the unit vectors in the direction of the $x$-axis and $y$-axis, respectively. The force constant of viscous drag force is:
\begin{equation}
    \mathbf{f}_h = -\eta_t v_{{h}}^{t} \mathbf{t} - \eta_p
    \left( \mathbf{v}_{{h}} - v_{{h}}^{t} \mathbf{t} \right),
\end{equation}
and its component along $y$-axis is
\begin{equation}
q_{h} =  \left( \eta_p - \eta_t \right) v_{{h}} \cos^2 \theta - \eta_p v_{{h}}
    \label{eq:Fp_withHead_withoutApprox}.
\end{equation}

Adding $q_h$ to Eq.~\ref{eq:longitudinalForce}, we get the final viscous force

\begin{equation}
\begin{aligned}
q =  \left( \eta_{p} - \eta_{t} \right) 
    \left( \omega_{{t}} R_{\textrm{d}} \sin \theta \cos \theta + v_{{h}} \cos^2 \theta \right) \\
    - \eta_p v_{{h}}.
    \label{eq:FpPerLength_full}
\end{aligned}
\end{equation}

We integrate this across the length of the flagella to compute the total propulsive force. Its component along the $x$-axis is

\begin{equation}
p_{h} = \left(\eta_{t} - \eta_{p} \right) v_{{h}} \sin \theta \cos \theta.
    \label{eq:forceHeadVel}
\end{equation}

Adding this solution to Eq.~\ref{eq:beamBending_withoutHead}, we get the following expression for the force constant along $x$-axis:

% \begin{align}
%     p &= p_{\textrm{t}} + p_{\textrm{h}} = EI \frac{\mathrm{d}^4 w}{\mathrm d y^4}, \\
%     p &=  \left(\eta_{t} - \eta_{p} \right) \left( v_{\textrm{h}} \sin^2 \theta + v_{\textrm{h}} \sin \theta \cos \theta \right) + \eta_{p}  v_{\textrm{h}}.
%     \label{eq:force_withHead_withoutApprox}
% \end{align}
%
\begin{align}
    p &= p_{t} + p_{h} = EI \frac{\mathrm{d}^4 w}{\mathrm d y^4}, \\
    % p &=  EI \frac{\mathrm{d}^4 w}{\mathrm d y^4} \\
    &= \eta_{p} v +  \frac{ \left(\eta_{t} - \eta_{p}\right)\left( v \left(\frac{\mathrm d w}{\mathrm d y}\right)^2 + v_{{h}}\left( \frac{  \mathrm d w}{\mathrm d y}\right) \right)} {\left( \frac{  \mathrm d w}{\mathrm d y}\right)^2 + 1}.
    \label{eq:force_withHead_withoutApprox}
\end{align}
%
% Once again, we integrate this expression to find the force in the $x$-direction. \continue
We used the $\texttt{bvpfcn}$ function in MATLAB to solve this nonlinear fourth order differential equations with boundary conditions
\begin{equation}
    w(0)=0, w' (0)=0, w''(L)=0, w'''(L)=0,
    \label{eq:BC}
\end{equation}
where the prime $(\;')$ indicates differentiation with respect to $y$. Since the head-induced effect is taken into account and no approximations about the flagellar deflection are made during the calculation, this design is referred to as \textit{nonlinear beam (NLB)} regime. However, by making some simplifying assumptions, we can achieve a more straightforward, closed-form solution. If we assume that the flagellum is slightly deflected to be treated as an Euler-Bernoulli beam, then $\sin^2 \theta \approx 0$ and $\cos^2 \theta \approx 1$. This enables us to simplify the solution and achieve linear approximation, referred to as \textit{linear beam (LB)} regime. 
The curvature in Euler-Bernoulli beam theory is approximated as $\frac{\mathrm{d}^2 w}{\mathrm d y^2}$ and $\left( \frac{\mathrm{d} w}{\mathrm d y} \right)^2$ terms are assumed to be negligible compared with 1. The boundary conditions on the beam are Eq.~\ref{eq:BC}. The deflection of a \textit{LB} is, therefore, calculated as follows 
\begin{equation}
    w(y) = \frac{p_ty^2(6L^2-4Ly+y^2)}{24EI},
\label{eq:deflection}
\end{equation}
and $p_t$ here is obtained as follows by omitting the $\sin^2 \theta$ term in Eq.~\ref{eq:beamBending_withoutHead}:
\begin{equation}
p_{t} = \eta_{p} v = \eta_{p} \omega_{{t}} R_{\textrm{d}}.
\label{eq:linearunitFx}
\end{equation}
Hence, the simplified total force along $x$-axis is
\begin{equation}
F_x = \eta_{p} \omega_{{t}} R_{\textrm{d}}L.
    \label{eq:linearFx}
\end{equation}
The linear propulsive force is likewise simplified as 
\begin{equation}
    q = \left( \eta_{p} - \eta_{t} \right) \omega_{{t}} R_{\textrm{d}} w'(y) - \eta_t v_{{h}},
    \label{eq:Fp_linear_unit}
\end{equation}
and the total propulsive force is 
\begin{equation}
    F_p = \left( \eta_{p} - \eta_{t} \right) \omega_{{t}} R_{\textrm{d}} w(L) - \eta_t v_{{h}} L.
    \label{eq:Fp_linear_solution}
\end{equation}
Note that we can neglect the head velocity in these equations because it has been experimentally proved to be small.
The deflection of \textit{LB}, \textit{NLB}, and \textit{NLB w/o head (nonlinear beam without head)} is compared in Fig~\ref{fig:comparison} and Section~\ref{subsec:deflection}. In addition, the head experiences a resistive force due to its translation and rotation. The viscous drags due to its translation and rotation are
\begin{equation}
\textbf{F}_p = -  C_1 (6\pi \mu R_\mathrm h \mathbf v_ h),
\label{eq:stokes_drag_refined}
\end{equation}

\begin{equation}
T_p = - C_2(8\pi \mu R_\mathrm h^3 \omega_ h),
\label{eq:stokes_torque_refined}
\end{equation}
respectively, where $6\pi \mu R_\mathrm h \mathbf v_h$ is the drag force and $8\pi \mu R_\mathrm h^3 \omega_h$ is the torque on a perfectly spherical object in a low Reynolds number fluid according to Stoke\rq{}s law and $\mathbf {v}_h = \dot{\mathbf x}_1$ is the velocity of the head, and $\omega_h = \Dot{\theta}^0$ is the rotational speed of the head. The coefficients $C_1$ and $C_2$ are included because our robot head is not a sphere; given the difficulty of measuring $C_1$, $C_2$, and $\mu$ during experiments, they are used as fitting parameters (see Section ~\ref{sec:paramFitting}) to match the experimental and simulation results in Section~\ref{sec:results}.
Furthermore, this flagellated robot system is balanced in terms of both force and torque. As a result, the propulsive force generated by the flagella should be equal to the drag force on the head, $F_p = \| \textbf{F}_p \|$, and the torque actuating the rotational movement of $n$ flagella should also be equal to the moment on the head. Using $F_p = \int_{y=0}^{y=L} q(y) \mathrm{d} y$, or the corresponding linear approximation in Eq.~\ref{eq:Fp_linear_solution}, we can compute the velocity:

\begin{equation}
    v_{{h}} = \frac{n F_p} {C_1 6 \pi \mu R_{\textrm{h}}},
    \label{eq:vHead}
\end{equation}
where $R_{\textrm{h}}$ is the radius of the robot head.
% The constant $C_1$ is used as a numerical pre-factor to account for the fact that the head is not entirely spherical.
%Combine Eq.~\ref{eq:vHead} and Eqs.~\ref{eq:deflection},~\ref{eq:linearunitFx}, ~\ref{eq:Fp_linear_solution}, 
%
Next, we calculate the rotation speed of the head, $\omega_{\textrm{h}}$, based on the torque balance. To compute this, we need the torque generated by $F_x$, which is given by 
\begin{equation}
    T_p = n R_{\textrm{d}} F_x,
    \label{Eq:Tp_flagella}
\end{equation}
where $F_x =\int_{y=0}^{y=L} p_t(y) \mathrm d y$, or with the linear approximation from Eq.~\ref{eq:linearFx}. Given that the torque generated by the flagella (in Eq.~\ref{Eq:Tp_flagella}) must be equal and in the opposite direction of the torque on the head in Eq.~\ref{eq:stokes_torque_refined},
%
% \begin{equation}
%     \omega_{\textrm{h}} C_2 8 \pi \mu R^3_{\textrm{h}} = n R_{\textrm{d}} F_x, \label{eq:torque_Stokes}
% \end{equation}
    we can compute the rotational speed of the head:
\begin{equation}
 \omega_{{h}} = \frac{n R_{\textrm{d}} F_x} {C_2 8 \pi \mu R^3_{\textrm{h}}}.
    \label{eq:OHead}
\end{equation}
%
% where the torque applied to the head (the left side of Eq.~\ref{eq:torque_Stokes}) is obtained according to Stoke\rq{}s law and $C_2$ is another numerical pre-factor to account for the non-spherical shape of the head. 
%
Plug Eq.~\ref{eq:sin_theta} into  Eq.~\ref{eq:beamBending_withoutHead}, we obtain $p_t$ as a function of only $y$ and $\omega_ t$, denoted as $p_t = f(y, \omega_t)$ and thus $F_x = f(\omega_t)$. Then, from Eq.~\ref{eq:OHead}, we find that $\omega_ h$ is a function of $\omega_ t$ whereas the remaining parameters are constants.
 In addition, one of control parameters in experiments is the rotational speed of the motor $\omega_{\textrm{T}}$:
% One outstanding issue is that in experiments, the control parameter is usually the rotational speed of the motor $\omega_{\textrm{T}}$, but the previous equations require $\omega_{\textrm{t}}$ to be known. We can express the rotational speed of the motor using this equation:
%
% \begin{equation}
%  \omega_{\textrm{T}} = \frac{N_f R_{\textrm{d}} F_x} {C_2 8 \pi \mu R^3_{\textrm{h}}} + \omega_{\textrm{t}}.
%     \label{eq:totalOmega}
% \end{equation}
%
\begin{equation}
 \omega_{{T}} = \omega_{{h}} + \omega_{{t}}.
    \label{eq:totalOmega}
\end{equation}
%
% We can begin with an approximated value of $\omega_{\textrm{t}}$ and use this equation to iteratively improve the accuracy of our guess as we test it.
%
\begin{figure*}[t]
\centering
\includegraphics[width=0.7\linewidth]{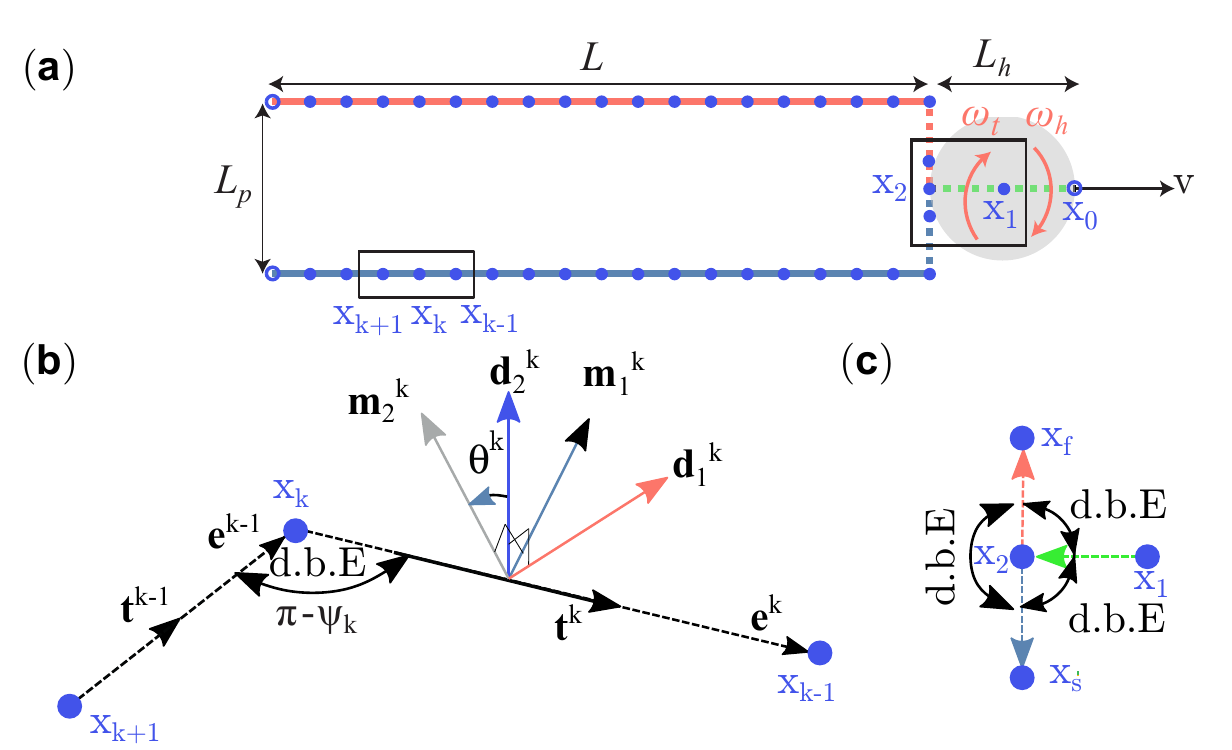}
\caption{
Discrete schematic diagram of a robot with two flagella. 
(a) Geometry of the robot in its undeformed state. 
% Where,  the solid blue line denotes the `neck' of the robot, i.e. the motor shaft reaching out of the head and the upper solid green line and bottom solid yellow line stand for two tails. 
Here, $L_h = 2R_\mathrm h$ is the diameter of the robot head, $L_p$ is the diameter of the 3D-printed plate bridging the head and flagella, and $L$ is the length of each flagellum. Dashed lines represent the rigid structure, while solid lines represent the elastic structure. The node $\mathbf x_1$ specifies the center of the head, i.e. its location. 
(\textbf{b}) A general close-up view of three adjacent nodes, $\mathbf x_{k-1}, \mathbf x_k,$ and $\mathbf x_{k+1}$, and two edges, $\mathbf e^{k-1} = \mathbf x_k - \mathbf x_{k-1}$ and $\mathbf e^{k} = \mathbf x_{k+1} - \mathbf x_{k}$. The turning angle, $\psi_k$, between the two edges results in bending energy and the rotation of the material frame from one edge to the next results in twisting energy. The reference frame on $\mathbf e^k$ is $ \left\{ \mathbf{d}_{1}^{k}, \mathbf{d}_{2}^{k}, \mathbf{t}^{k} \right\} $ and the material frame is $ \left\{ \mathbf{m}_{1}^{k}, \mathbf{m}_{2}^{k}, \mathbf{t}^{k} \right\} $. The twist angle from the first material director associated with $\mathbf{t}^{k-1}$ to the next material director associated with $\mathbf{t}^{k}$ is $\theta^k$.
(c) A close-up of the ``T"-shape joint node $\mathbf x_2$ that connects the head to the flagella. d.B.E. indicates discrete bending and twisting energy. Only the joint node, $\mathbf x_2$, is connected to more than two nodes.}
\label{fig:simSchematic}
\end{figure*}
Note that $\omega_h$ and $\omega_t$ here are scalars, but the rotational directions of the head and flagella of the robot are opposite.
In conclusion, we have two equations for two unknown variables $\omega_t$ and $\omega_h$, allowing us to solve the propulsion of the entire system. From these two variables, all other parameters, such as $v_ h$ and $F_p$ (by combining Eqs.~\ref{eq:Fp_linear_solution} and ~\ref{eq:vHead}) can be calculated. We emphasize that the overall framework includes not only the flagellar propulsion based on beam theory but also the head motion based on Stokes\rq{}law.
\subsection{Numerical method of flagella locomotion}
\label{sec:NumerialFramework}
In this section, we present the numerical simulation framework in which the robot structure is modeled as a network of connected Kirchhoff elastic rods. The elastic energies of this structure are thereby the linear sum of bending, twisting, and stretching energies, the negative gradient of which is the elastic forces on each degree of freedom (DOF). The implicit Euler method is used to solve equations of motion (EOM) in which the external drag forces from the GM on each DOF are modeled through RFT. When $\omega_T$ is small, locomotion parameters obtained from this simulation, e.g., velocity, propulsive force, and deflection, are compared to those obtained from analytical equations based on Euler-Bernoulli beam theory described in Section~\ref{sec:beam}. Furthermore, the simulation results of DDG are shown to match quantitatively with experimental data, validating the method itself.

\subsubsection{Discrete Differential Geometry (DDG)}
\label{sec:DER} 
% \textit{Kinematics} 
% \fix{A deeper subsection is needed}
% \label{sec:kinematics}
Fig~\ref{fig:simSchematic} is a discrete representation of a robot with a rigid head and 3D-printed plate and $n$ straight elastic flagella ($n=2$ in the figure). The numerical values for physical parameters will be given in Section~\ref{subsec:parameters}. Along the schematic discretization, there are a collection of $N$ nodes (circles in Fig.~\ref{fig:simSchematic}), $\mathbf x_0, \mathbf x_1, ..., \mathbf x_{N-1} $, attached via $N-1$ thin elastic rod vectors, $\mathbf e^k = \mathbf x_{k+1}-\mathbf x_k (k=0, 1, ..., N-2)$, called edges (the lines between two adjacent nodes). Hereafter, we use subscripts for node-based quantities and superscripts for edge-based quantities. As shown in Fig.~\ref{fig:simSchematic}(b), the head is discretized into $\mathbf x_0, \mathbf x_1,\mathbf x_2$, with node $\mathbf x_2$ denoting the one at the head-flagella junction. It is a unique node that is connected to $n+1$ additional nodes. All other nodes are connected to two nodes, or to a single node in case of the last node at the end of each flagellum, denoted by an open circle.

Rather than being just a single vector, each edge (see Fig.~\ref{fig:simSchematic}) is also equipped with two corresponding sets of orthonormal reference frames to track its rotation, a reference frame  $ \left\{ \mathbf{d}_{1}^{k}, \mathbf{d}_{2}^{k}, \mathbf{t}^{k} \right\} $ and a material frame $ \left\{ \mathbf{m}_{1}^{k}, \mathbf{m}_{2}^{k}, \mathbf{t}^{k} \right\} $; both of them share the tangent $\mathbf{t}^{k} = \mathbf{e}^{k} / \| \mathbf{e}^{k} \|$ as one of the directors ($\| \cdot \|$ represents the Euclidean norm of a vector). The reference frame serves as a frame initialized at time $t=0$ and updated at each subsequent time step via time-parallel transport, as illustrated in Fig.~\ref{fig:simSchematic}(c), the material frame can be calculated in terms of a scalar twist angle, $\theta^k$. The detailed transformation expression between the reference and material frames can be found in Du \textit{et al.}\cite{du2021modeling} and Jawed \textit{et al.}\cite{jawed2018primer}. For this class of flagellated robots, the DOF vector is constituted of node positions and twist angles formulated as follows
\begin{equation}
\mathbf q = \left[
\mathbf x_0, \mathbf x_1, \mathbf x_2, \ldots, \mathbf x_{N-1}, \theta^0, \theta^1, \ldots, \theta^{N_e-1} \right]^T,
\end{equation}
where $N_e$ is the total number of edges in the entire robot, which in this work is $N-1$ (see Fig.~\ref{fig:simSchematic}(a)), and the superscript $^T$ stands for transpose. $\mathbf q$ has a dimension of  $\textrm{ndof}=3\times N + N_e$. Since $\mathbf q$ completely defines the configuration of the robot whose deformation varies with time, the DOF vector is a function of time, i.e. $\mathbf q \equiv \mathbf q (t)$. At time $t=0$, the robot is undeformed and the DOF vector is $\mathbf q(0) \equiv \bar{\mathbf q}$; in the following, $\bar{(\; )}$ represents evaluation of a quantity in the undeformed configuration. The strain calculations coming next all rely on the DOF vector $\mathbf q$.
%
% \begin{subequations}
% \begin{align}
% \mathbf{m}_{1}^{k} &= \mathbf{d}_{1}^{k} \cos \theta^{k} + \mathbf{d}_{2}^{k} \sin \theta^{k} \label{MaterialFrame1}\\
% \mathbf{m}_{2}^{k} &= - \mathbf{d}_{1}^{k} \sin \theta^{k} + \mathbf{d}_{2}^{k} \cos \theta^{k}
% \label{MaterialFrame2}
% \end{align}
% \label{eq:materialFrame}
% \end{subequations}
%
Based on this kinematic representation, we will sequentially discuss the formulation of elastic energies and forces, external forces, and simulation loops. 

\subsubsection{Elastic energies and forces}
\label{sec:elasticenergy}
The total elastic energy of a flagellated robot structure is the linear sum of stretching, bending, and twisting energies:
\begin{equation}
E_\textrm{E} = E_s + E_b + E_t,
\label{eq:elasticEnergy}
\end{equation}
where $E_s, E_b,$ and $E_t$ are the stretching, bending, and twisting energies, respectively.

Elastic energies of a structure are associated with the corresponding macroscopic strains, axial stretch, curvature, and twist\cite{audoly2000elasticity}. The stretching energy associated with each edge is related to the axial stretch of the edge. Axial stretch is the change in length of an edge, normalized by the undeformed length. The axial stretch,  $\epsilon^{k}$, of edge $e^k$ is
\begin{equation}
\epsilon^{k} = \frac { \| \mathbf{e}^{k} \| } { \| \bar{\mathbf{e}}^{k} \| } - 1,
\label{eq:stretch}
\end{equation}
where $\| \bar{\mathbf{e}}^{k} \|$ is the undeformed edge length. Given this axial stretch, the stretching energy along edge $e^k$ is 
\begin{equation}
E_s^k = \frac{1}{2} EA \left( \epsilon^{k} \right)^2 \| \bar{\mathbf{e}}^{k} \|
\label{eq:stretchingEnergy}
\end{equation}
where $E$ is the Young\rq{}s modulus, $A=\pi r_0^2$ is the cross-sectional area of the flagella, and $r_0$ is the flagellar radius. The total stretching energy of the robot is the sum of individual stretching energies, i.e.
\begin{equation}
E_s =  \sum_{k=1}^{N_e} E_s^{k}.	
\label{eq:totalstretchingEnergy}
\end{equation}
For edges on the rigid head and disk, the stretching stiffness $EA$ is set to be sufficiently large to ensure that deformation is negligible.

Bending energy is related to curvature, a node-based quantity that is related to the turning angle, $\psi_k$, displayed in Fig.~\ref{fig:simSchematic}(c). Note that curvature is applied to all nodes except terminal nodes, and the curvature binomial for node $x_k$ is $(\mathbf{\kappa b})_{k}$ defined as the following vector 
\begin{equation}
(\mathbf{\kappa b})_{k} = \frac {2 \mathbf{e}^{k-1} \times \mathbf{e}^{k} } { \| \mathbf{e}^{k-1} \| \| \mathbf{e}^{k} \| + \mathbf{e}^{k-1} \cdot \mathbf{e}^{k} },
\label{curvature}
\end{equation}
where $\| (\mathbf{\kappa b})_{k} \|  = 2 \tan \left( \frac{\psi_k}{2} \right)$.
The scalar curvatures along the first and second material directors, calculated using the curvature binomial, are
\begin{subequations}
\begin{align}
\kappa_{k}^{(1)} &= \frac {1} {2} (\mathbf{m}_{2}^{k-1} + \mathbf{m}_{2}^{k}) \cdot (\mathbf{\kappa b})_{k}, \label{BendingCurvature1} \\
\kappa_{k}^{(2)} &= \frac {1} {2} (\mathbf{m}_{1}^{k-1} + \mathbf{m}_{1}^{k}) \cdot (\mathbf{\kappa b})_{k} \label{BendingCurvature2}.
\end{align}
\end{subequations}
The bending energy is then calculated according to the equation 
\begin{equation}
E_{\text{b}} = \sum \frac {1} {2} \frac {EI} {\Delta l_{k}} \left[ \left( \kappa_{k}^{(1)} - \bar{\kappa}_{k}^{(1)} \right)^2 + \left( \kappa_{k}^{(2)} - \bar{\kappa}_{k}^{(2)} \right)^2 \right],
\label{eq:bendingEnergy}
\end{equation}
where $\sum$ indicates summation over all the discrete bending energies, $ \Delta l_{k} = \frac {1}{2} \left( \| \bar{\mathbf{e}}^{k-1} \| + \| \bar{\mathbf{e}}^{k} \| \right) $ is the Voronoi length for the $k$-th node, $ \bar{\kappa}_{k}^{(1)} $ and $ \bar{\kappa}_{k}^{(2)} $ are the material curvatures in the undeformed configuration, and $EI = \frac{\pi}{4}Er_0^4$ is the bending stiffness of the rod. For the rigid robotic head and 3D-printed plate, the value of bending stiffness $EI$ is assumed to be so large that the curvatures at the rigid nodes remain nearly constant throughout the simulation.

Finally, the twisting energy is related to the relative rotation of the material frames between two adjacent edges, i.e. twist. The twist at the $k$-th node is 
\begin{equation}
\tau_{k} = \theta^k - \theta^{k-1} + \Delta m_{k, \textrm{ref}},
\end{equation}
where $\Delta m_{k, \textrm{ref}}$ is the reference twist, which is the twist of the reference frame as it moves from the $(k-1)$-th edge to the $k$-th edge. 
The method by which we calculate this reference twist is detailed at the end of Section 4.2 in our previous work\cite{du2021modeling}. 
% We calculate this reference twist as follows. The first reference frame director, $ \mathbf{d}_{1}^{k-1}$, is parallel transported from the $(k-1)$-th edge to the $k$-th edge to find $\mathbf{d}_\textrm{tmp}$. Parallel transporting the reference director is done by moving it from one edge to the next without twisting according to the following steps:
% %
% \begin{eqnarray*}
% & \mathbf b &= \mathbf t^{k-1} \times \mathbf t^k,\\
% & \hat {\mathbf b} &= \frac{\mathbf b}{\|\mathbf b \|},\\
% & \mathbf n_1 &= \mathbf t^{k-1} \times \hat {\mathbf b},\\
% & \mathbf n_2 &= \mathbf t^k \times \hat {\mathbf b},\\
% & \mathbf{d}_\textrm{tmp} &= (\mathbf{d}_{1}^{k-1} \cdot \mathbf t^{k-1}) \mathbf t^k + (\mathbf{d}_{1}^{k-1} \cdot \mathbf n_1) \mathbf n_2 + (\mathbf{d}_{1}^{k-1} \cdot \hat {\mathbf b}) \hat {\mathbf b},
% \end{eqnarray*}
% where $\mathbf t^{k-1}$ and $\mathbf t^k$ are the tangents on the $(k-1)$-th and $k$-th edges, respectively. The reference twist, $\Delta m_{k, \textrm{ref}}$, is the signed angle from $\mathbf{d}_\textrm{tmp}$ to $\mathbf{d}_{1}^k$ about $\mathbf t^k$.

The twisting energy is then calculated according to the equation 
\begin{equation}
E_t = \sum \frac {1} {2} \frac {GJ} {\Delta l_{k}} \left( \tau_{k} - \bar{\tau_{k}} \right)^2,
\label{eq:twistingEnergy}
\end{equation}
where $ \bar{\tau_{k}} $ is the undeformed twist along the centerline, $G$ is the shear modulus, and $GJ =  \frac{\pi}{2}Gr_0^2$ is the twisting stiffness. For the rigid components, the twisting stiffness is sufficiently large. Additionally, we assume that the material of flagella is nearly incompressible (i.e. Poisson\rq{}s ratio $\nu=0.5$), so $G = E/3$.

In summary, at each DOF $\mathbf q_k$, the elastic forces (affiliated with nodal position) and elastic moments (affiliated with the twist angles) are
\begin{equation}
    F_k^{\textrm{E}} = -\frac{\partial E_\textrm{E}}{\partial \mathbf q_k},
\end{equation}
where $k = 0, 1, ..., \textrm{ndof}-1$.

In a single elastic rod, each internal node is associated with a discrete bending and twisting energy. However, in this paper, the flagellated robot is represented as a network of rods and a ``T"-shape joint node. This node is associated with multiple discrete bending and twisting energies(denoted as d.B.E in Fig.~\ref{fig:simSchematic}(b)). In order to simulate the dynamics of the robot, EOM for each DOF is required, which includes not only elastic forces but also external forces. Consequently, we illustrate how we model the external forces from the GM onto the robot in the next section.
% \subsubsection{External forces}
% \label{sec:externalforces}
% The external viscous force exerted by the GM on the flagella is modeled using RFT, with details provided in Eqs.~\ref{eq:externalForceTail}-~\ref{eq:viscousForce} in Section~\ref{subsubsec:flagellarPropulsion}. Only the subscript $k$ needs to be added to denote the external force at $k$-th node. 

\subsubsection{Fully-implicit simulation}
\label{sec:discretesimulation}
In order to simulate the locomotion of the robot, time is discretized into small time steps of size $\Delta t$. At each time step $t_j$, the DOF vector $\mathbf{q}$ is updated. Using the following equations of motion (EOM), the $k$-th DOF marches from $t=t_j$ to $t=t_{j+1}=t_j + \Delta t$:

\begin{equation}
\frac{m_k} {\Delta t} \left[
\frac{ q_k (t_{j+1}) - q_k (t_j) } { \Delta t } - 
\dot{q}_k (t_j) \right] - F_k^{\textrm{E}} -
F_k^{\textrm{ext}} = 0,
\label{eq:DER3D_EOM}
\end{equation}
where $q_k(t_j)$ and $\dot{q}_k(t_j)$ are the known DOF and velocities at the previous time step, respectively, $E_\textrm{E}$ is the elastic energy evaluated at $t_{j+1}$, $F_k^\mathrm {ext}$ is the external force (or torque for twist angles) on the $k$-th DOF, and $m_k$ is the lumped mass at the DOF. The external viscous force exerted by the GM onto the flagella are detailed as Eqs.~\ref{eq:externalForceTail}-~\ref{eq:viscousForce} in Section~\ref{subsubsec:flagellarPropulsion}.The external viscous force and torque onto the head are calculated through Eqs.~\ref{eq:stokes_drag_refined} and ~\ref{eq:stokes_torque_refined}.  Only the subscript $k$ needs to be added to denote the force/torque at $k$-th node. Because the dynamics of this system is dominated by viscous forces with negligible influence of inertia, the results presented are not mass dependent as long as low Reynolds number is maintained. Eq.~\ref{eq:DER3D_EOM} represents the collection of $\textrm{ndof}$ equations that has to be solved to get the new DOF $q_k(t_{j+1})$. Essentially, this equation is a statement of ``mass times acceleration = elastic force + external force" at the $k$-th DOF. Once $q_k(t_{j+1})$ is updated, the velocity at time $t_{j+1}$ is determined as $\dot{q}_k (t_{j+1}) = \left( q_k(t_{j+1}) - q_k(t_{j}) \right) / \Delta t$.

The EOM are solved using the Newton-Raphson method, i.e. 
\begin{equation}
    \mathbb J \Delta \mathbf q = \mathbf f,
\label{eq:Newton-Raphson}
\end{equation}
where $\mathbf f$ is a vector of size $\textrm{ndof}$, the $k$-th component of this vector can be computed from Eq.~\ref{eq:DER3D_EOM}, and $\mathbb J$ is a square matrix representing the Jacobian for Eq.~\ref{eq:DER3D_EOM}. The $(k,i)$-th element in the Jacobian matrix is
\begin{equation}
\mathbb J_{ki} = \frac{\partial f_k}{\partial \zeta_i} = \frac{m_k}{\Delta t^2} \delta_{ki} + \frac{\partial^2 E_{\textrm{E}}}{\partial q_k \partial q_i} - \frac{\partial F_k^{\textrm{ext}}}{\partial q_i},
\label{eq:DER3D_Jacobian}
\end{equation}
where $\delta_{ki}$ represents Kronecker delta, the terms are gradient of inertia, elastic forces, and external forces, respectively, in the order shown in Fig.~\ref{eq:DER3D_EOM}. Well-documented evaluation of the gradient of the elastic energy ($\frac{\partial E_{\textrm{E}}}{\partial q_k}$) as well as its Hessian ($\frac{\partial^2 E_{\textrm{E}}}{\partial q_k \partial q_i}$) can be found in Jawed \textit{et al.}\cite{jawed2018primer} and Bergou \textit{et al.}\cite{bergou2010discrete}. 
\subsubsection{Remarks on algorithm}
\label{sec:algoUniqueness}
Next, we summarize the novelty of our algorithm for simulating a robot with multiple flagella described above. Note that solving  Eq.~\ref{eq:Newton-Raphson} is the most computationally expensive part of the entire simulation procedure. It is crucial to notice the sparsity of the Jacobian matrix, $\mathbb J$, and exploit its sparsity during the solution process\cite{schenk2002solving}, which helps reduce the computation cost. If the structure to be simulated is a single elastic rod (unlike a network of rods in this paper), the Jacobian is banded, and the time complexity of this algorithm is $O(N)$\cite{bergou2010discrete}. However, the Jacobian in this paper is not banded due to the presence of the joint node $\mathbf x_2$ in Figs.~\ref{fig:simSchematic}(\textbf{a})(\textbf{c}).
Referring to Fig.~\ref{fig:simSchematic}(\textbf{b}), the entire structure is a combination of stretching springs (e.g. one stretching spring is between $\mathbf x_k$ and $\mathbf x_{k+1}$) and bending-twisting springs (e.g. one bending-twisting spring is between $\mathbf x_{k-1}, \mathbf x_k,$ and $\mathbf x_{k+1}$). The stretching energy of each spring (Eq.~\ref{eq:stretchingEnergy}) only depends on six DOFs (nodal coordinates of two nodes). For the stretching spring on edge $\mathbf e^k$, these DOFs are $\mathbf x_k$ and $\mathbf x_{k+1}$. The gradient vector $\left( \frac{\partial}{\partial \mathbf q} \left[ \frac{1}{2} EA \left( \epsilon^{k} \right)^2 \| \bar{\mathbf{e}}^{k} \| \right] \right)$ has only six non-zero terms and the Hessian matrix $\left( \frac{\partial^2}{\partial \mathbf q \partial \mathbf q} \left[ \frac{1}{2} EA \left( \epsilon^{k} \right)^2 \| \bar{\mathbf{e}}^{k} \| \right] \right)$ has only $6 \times 6$ non-zero terms.
As for the bending and the twisting energies of each spring (Eqs.~\ref{eq:bendingEnergy} - \ref{eq:twistingEnergy}), they are only dependent on eleven DOFs, i.e. $\mathbf x_{k-1}, \theta^{k-1}, \mathbf x_k, \theta^k,$ and $\mathbf x_{k+1}$ in case of the spring at $\mathbf x_k$ in Fig.~\ref{fig:simSchematic}(\textbf{b}).
As a result, the gradient vector and Hessian matrix corresponding to these two energies contain only eleven and $11 \times 11$ non-zero terms, respectively. The complete expressions for the gradient and Hessian terms are available in works\cite{bergou2008discrete, jawed2018primer, panetta2019x}; the coding implementation can also be found in open-source repositories\cite{jawed2014coiling, panetta2019x, choi2021implicit}.

Last but not least, a crucial contribution of this study is the observation that the actuation of the robot, e.g., the rotation of the motor, can be readily accounted for in the framework above by updating the undeformed configurations with time. Normally, the undeformed configuration of a structure is fixed and assumed to remain constant throughout the simulation. The strains in undeformed configuration (e.g. $\bar{\kappa}_{k}^{(1)}, \bar{\kappa}_{k}^{(2)}, \bar \tau_k$ in Eqs.~\ref{BendingCurvature1}, ~\ref{BendingCurvature2}, and ~\ref{eq:twistingEnergy}) are used in calculation of elastic energies, their gradient (i.e. elastic forces), and Hessian. 
However, in this class of robots, the rotation of the motor causes the undeformed twist at the head node ($\mathbf x_1$) to vary with time. At each time step $t_j$, the robot is actuated by updating the undeformed twist of the first node according to the rotational speed of the motor, $\omega_T$:
\begin{equation}
\bar \tau_0 (t_j) = \omega_T t_{j}.
\label{eq:SpecifyTheta}
\end{equation}
This actuation causes the rotations of the head ($\omega_h$) and the flagella ($\omega_t$) along opposite directions such that $|\omega_T| = |\omega_h| + |\omega_t|$. The total rotational speed, $\omega_T$, is a control parameter in this study while the other one is the number of flagella, $n$.
\subsection{Physical parameters} 
\label{subsec:parameters}
The following session will illustrate how we vary the robot design and experimental setup to realize designs 1 and 2. Remember that design 1 corresponds to the phenomenon that the propulsive speed of the robot decreases as the number of flagella increases, whereas design 2 reverses this phenomenon. Because the robot designs utilized in these two designs are different, we will provide the geometric and material parameters of the robots during experiments as follows. Hereafter, if we do not specify which design the parameters apply to, they apply to both.
The number of flagella is $n=2$ or $3$ and each flagellum has a length of $L=0.111$ m (design 1) and $L=0.089$ m (design 2), a radius of $r_0 = 3.2$ mm, a density of $1000$kg/m$^3$ (which is used to compute $m_i$ in Eq.~\ref{eq:DER3D_EOM}), a Young\rq{} modulus $E = 1.2 \times 10^6$ N/m$^2$, and Poisson\rq{} ratio $\nu = 0.5$ (incompressible material). Referring to Fig.~\ref{fig:simSchematic}, the radius of the robot head is $R_\mathrm h=0.02$ m (design 1) and $R_\mathrm h=0.015$ m (design 2), and the diameter of 3D-printed circular plate is $L_p=0.04$ m (design 1) and $L_p=0.03$ m (design 2).
Recall that parameters $C_1$, $C_2$ and $\mu$ in Eqs.~\ref{eq:vHead} and ~\ref{eq:OHead} are fitting parameters and will be fitted later in Section~\ref{subsubsec:VelocityVSFlagellaNumber}. The time step used in this paper is $\Delta t=10^{-2}$ s and the length of each edge on flagella (in undeformed state) is $\| \bar {\mathbf e}^k \| = 4.11$ mm. Convergence studies were performed to ensure that the size of temporal and spatial discretization ($\Delta t, \| \bar {\mathbf e}^j \|$) has negligible impact on the simulation performance. 
%
% \subsubsection{Euler-Bernoulli beam simulation}
% \label{subsubsec:beamSim}
% When $\omega_T$ is relatively small($\lesssim 5 \mathrm{rpm}$), the flagellar deformation drops to the \textit{LB} design introduced in Section~\ref{subsubsec:headeffect}
% The input angular velocity, $\omega_T$, is 5 rpm, which is small enough that all solution methods should converge to the same results. For the DER simulation, we parametrize the flagella using 400 nodes to maximize the accuracy of the simulation.

% \begin{figure}[h!]
%     \centering
%     \includegraphics[width=0.75\textwidth]{figures/Fig_sim_schematic.pdf}
%     \caption{Schematic showing the forces and velocities on the flagellum.}
%     \label{fig:schematic}
% \end{figure}

\section{Experimental Design}
\label{sec:experiment}
\begin{figure}
    \centering
    \includegraphics[width=0.75\columnwidth]{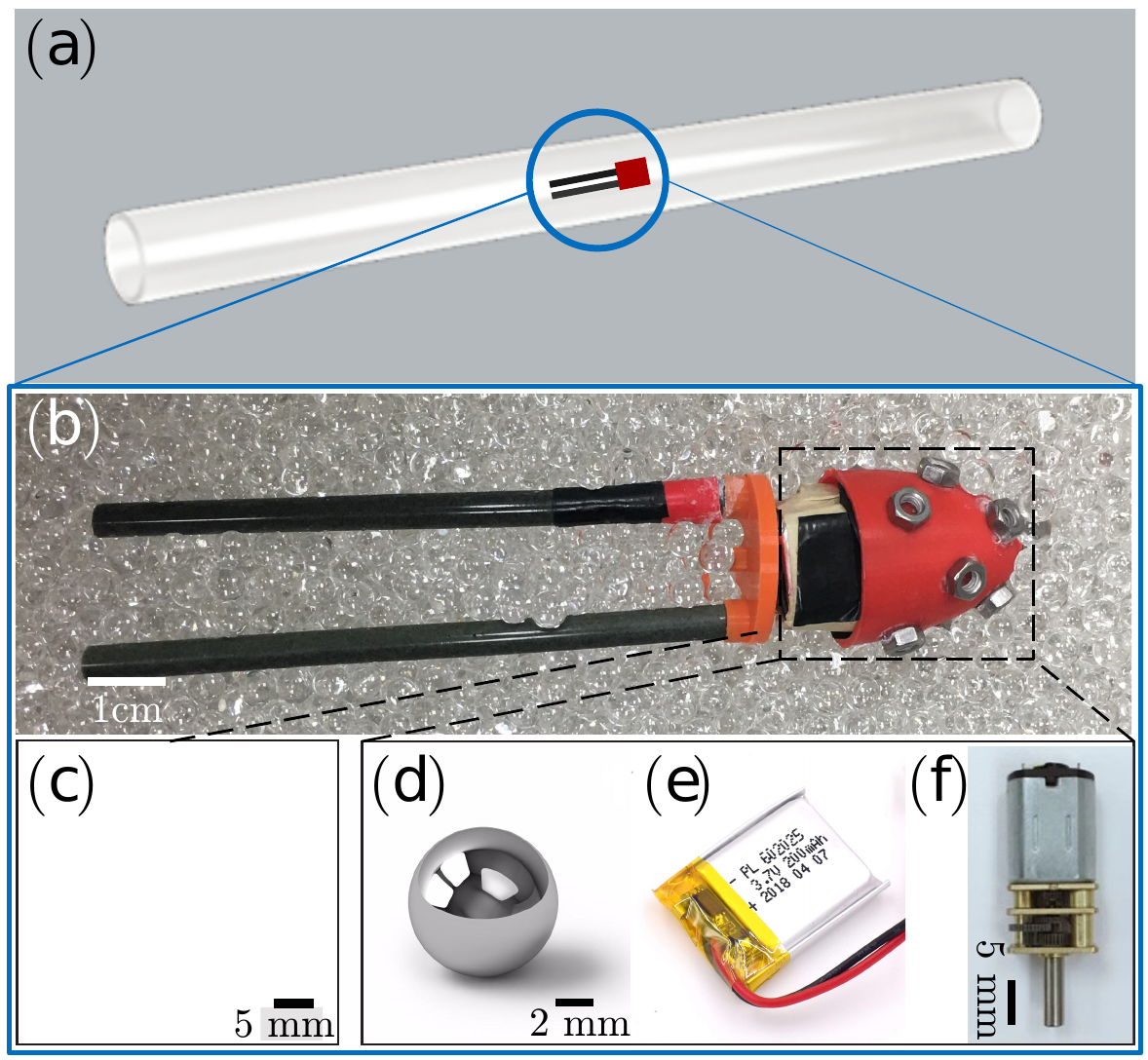}
    \caption{The compositive perspective of the experimental system for design 2. The one for design 1 can be found as Fig. 2 in our previous work\cite{du2021modeling}. Here, (\textbf{a}) the robot maneuvers inside the GM, which is contained inside a transparent cylindrical tube. (\textbf{b}) the robot is comprised of $n=2$ flagella and a conic head. (\textbf{c}) A circular 3d-printed plate bridges the flagella with the head. The head consists of three components: (\textbf{d}) chrome steel bearing balls, (\textbf{e}) batteries and (\textbf{f}) a motor.}
    \label{fig:exp_setup}
\end{figure}

As the flagellated locomotion is an intricate interplay between the elasticity of the flagella and hydrodynamic loading from the surrounding GM, our experiments considered both the robot design and selection of GM for propulsion. The robot is placed inside a transparent cylindrical tube filled with GM as shown in Fig.~\ref{fig:exp_setup} (\textbf{a}). Before finalizing the cylindrical tube as the container, we initially placed the robot at the midline of the height in a rectangular tank of 50$\times$50$\times$40cm or a circular tube with a radius of 25mm, but the robot moved closer and closer to the surface of the GM because of a drag-induced lift\cite{maladen2011undulatory} though the robot was manufactured as symmetrically as possible. \par

\subsection{Robot structure}
\label{subsec:robotdesign}
Given that Euler-Bernoulli beam theory in Section~\ref{sec:beam} predicts the presence of designs 1 and 2, experiments were performed to realize the prediction. Similar to the compact and lightweight robot used to realize design 1 (see Fig. 2 in our prior work\cite{du2021modeling}), the photograph in Fig.~\ref{fig:exp_setup}(\textbf{b}) depicts our robot for design 2 realization, which is propelled by $n$ soft elastic flagella. It consists of four distinct components: multiple straight elastic flagella, Fig.~\ref{fig:exp_setup}(\textbf{c}) one 3D-printed circular plate attached to the motor shaft that protrudes from the robot head to hold flagella, and a head embedded with Fig.~\ref{fig:exp_setup}(\textbf{(d)}) eight steel bearing balls (G25 Chrome Steel-AISI 52100) with an approximate diameter of 7.74 mm and mass of 2.06 grams, Fig.~\ref{fig:exp_setup}(\textbf{(e)}) two 3.7V 200mAh rechargeable 502025 LiPo batteries (from Du litter energy battery), and Fig.~\ref{fig:exp_setup}(\textbf{(f)}) one DC geared motor (from uxcell) with a nominal voltage of 3V, a power output of 0.35W, and a stall current of 0.55A. Inspired by natural creatures such as scorpions, snakes, and sand lizards, which have heads that are not spherical but more triangular, the head of our robot is designed as a cuboid (see Fig. 2 in Du \textit{et al.}) or nose cone (see Fig.~\ref{fig:exp_setup}(\textbf{b})).
% The former head shape is used in design 1 experiments, where the speed of the robot decreases as the number of flagella, $n$, increases while the latter is used in design 2 experiments, which is the inverse of the former. 
According to Euler-Bernoulli beam theory-based analyses, illustrated in Section~\ref{subsubsec:headeffect}, a smaller $\omega_h$ is advantageous for the realization of design 2. As a result, compared to the previous cuboid head design, the head in Fig.~\ref{fig:exp_setup}(\textbf{b})) has not only a different shape but also higher inertia and a more frictional surface. Metal balls in Fig.~\ref{fig:exp_setup}(\textbf{(d)} are added to increase the inertia. Both head shapes contribute to the fluidization of the GM in front of the robot. To fabricate the elastic flagella, we applied the molding and casting techniques developed by Lazarus \textit{et al.}\cite{lazarus2013contorting} and Miller \textit{et al.}\cite{miller2014shapes}. The silicone-based rubber (vinylpolysiloxane from Elite Zhermack) injected into a PVC tube (e.g. from VWR International) mold and the inner and outer diameters of the PVC tube are the same as in our previous work\cite{du2021modeling}, $3.175$ and $6.35$ mm. The radius of elastic flagella can be varied by using PVC tubes  with different inner and outer diameters, allowing our robot platform to be effortlessly scaled up or down. 
% The flagellar length $l$ varies from 49 \textrm{cm} to 109 \textrm{cm}, and $n$ is two or three. 
The relationship between $n$ and robot speed $v$ is investigated in detail and will be explained in Section~\ref{subsec:speedOfTheRobot}. \par 
The spinning soft flagella propel the robot at a rotational speed of $\omega_{{t}}$. Since the torque is system-balanced, the head is then actuated at a rotational speed of $-\omega_{{h}}$. The control parameter is the rotational speed of the flagella relative to the head, $\omega_{{T}} = \omega_{{t}} + \omega_{{h}}$. To vary $\omega_{{T}}$, we replaced the geared motors while maintaining the other components the same instead of adding an encoder, which would significantly increase the robot size. The size and weight of all the motors are almost the same, (15-17) $\times$ 12 $\times$ 10 mm (L$\times$W$\times$H) and 13-15g even if they supply different $\omega_h$. When necessary, we added electrical tapes around the motor to account for the small differences in size and weight for different motors. Moreover, the value of $\omega_t$ is proportional to the battery voltage and will decrease when the voltage drops. Subsequently, to maintain a constant $\omega_t$, we fully charged the batteries before each experiment and recharged them after each approximately ten-minute experimental trial.\par 

\subsection{Granular medium} 
\label{subsec:granular}
% Two simultaneous motions, the rotation of the robot’s head and tails and the translation of the robot’s body in the axial direction, are coupled during the movement. The rotational speeds of the robot’s head, $\omega_{\textrm{h}}$, and tails, $\omega_{\textrm{t}}$, are therefore needed to investigate the coupling.
A granular medium has to be chosen to complement the untethered flagellated robot design. Too much internal friction between granules prevents the robot head from fluidizing the granules in front and propelling forward, while too little friction results in insufficient flagellar propulsion. Internal friction during locomotion is correlated with the phase transition of GM from solid to fluid, which is controlled by temperature and volume fraction (VF)\cite{gravish2014effect}. The less effort required to alter the VF of GM, the more beneficial it is for studying the locomotion in GM.  
Apart from the reason mentioned above, we chose gel soil water crystal beads (from EBOOT) as the GM for their transparency. Robotic performance, e.g., the robot position and the rotational speeds of head and flagella, could be viewed outside the medium using a traditional digital camera (Nikon D3400). The beads had a diameter of $2.5$ mm when dried and $d_f = 9.4 \pm 0.4$ mm when completely saturated with water. The size of beads is controllable by adjusting the time they are submerged in water and is reversible after dehydration. During experiments, the robot for design 1 ran in the GM with $d_f = 9.4 \pm 0.4$ mm whereas the one for design 2 ran in the GM with $d_f = 5.2 \pm 0.4$ mm (see Fig.~\ref{fig:exp_setup}(b)). The corresponding VF values were approximately $0.52$ and $0.54$. In the former case, $v$ decreases as $n$ decreases (design 1), while in the latter case, the converse (design 2) is true. Throughout all experiments in designs 1 or 2, the temperature variation of GM is kept to a maximum of 0.5 degrees, and the granule configuration, including size, density, and homogeneity, is maintained.

\subsection{Locomotion in granular medium}
\label{subsec:LocomotionExp}
We used a transparent acrylic cylindrical tube (from FixtureDisplays, Amazon) with an inner radius of $53$ mm and an axial length of $1220$ mm as the reservoir for the GM to conduct the locomotion experiments. The tube was filled with GM and placed horizontally, perpendicular to the direction of gravity. Before each experiment trial, we placed the robot at one end at the center of the cross-section of the tube. Since the force applied by the GM on the robot in the cylindrical tube is axially symmetric, the confinement effect of the relatively close bounding wall is canceled out. As the robot was positioned at the center and thus surrounded by compact granules, the drag-induced lift\cite{maladen2011undulatory} was suppressed. Hence, the robot actuated by the rotating flagella would move in GM along a roughly straight line. The video camera captured this movement at a frame rate of 29.98 fps. In addition, both the head and flagella were marked with markers of a different color than the corresponding robot components. For instance, as shown in Fig.~\ref{fig:exp_setup}(\textbf{b}), a black marker was attached to the yellow-colored head, and a red marker was attached to one of the dark green colored elastic flagella. Together with the GM's transparency, this operation ensured the accuracy of counting the rotational speeds of the robot head ($\omega_h$) and flagella ($\omega_t$) and locating the position of the robot, $s$ in recorded experimental videos. 

\section{Results and discussion}
\label{sec:results}
Recall that there are two control parameters in our study, the total rotational speed of the motor embedded in the head, $\omega_T$, and the number of flagella, $n$. As shown in Fig.~\ref{fig:simulationSnapshots}, when the motor is powered on, the actuation from the motor, $\omega_T$, is split into a constant rotational speed of the head, $\omega_h$, and flagella, $\omega_t$, and the relationship $\omega_T = \omega_h + \omega_t$ is always satisfied ($\omega_T, \omega_h,$ and $\omega_t$ are all non-negative values). Because the system is torque-balanced, the head and flagella rotate in opposite directions. The net propulsive force is the residual of the propulsion generated by flagellar deformation (with the deflection $w$) and the drag on the head and flagella from the GM, propelling the entire robot forward (along with $-y$ direction) at speed $v$ as illustrated in Fig.~\ref{fig:schematic}.
In this section, we display the result comparison among the experiments and beam theory-based and DER-based numerical simulations. Since there are two designs, the sections below cover two sets of fitting parameters, experimental data, and numerical results.

\subsection{Speed of the robot}
\label{subsec:speedOfTheRobot}
\begin{figure*}[t!]
    \centering
    \includegraphics[width=0.75\textwidth]{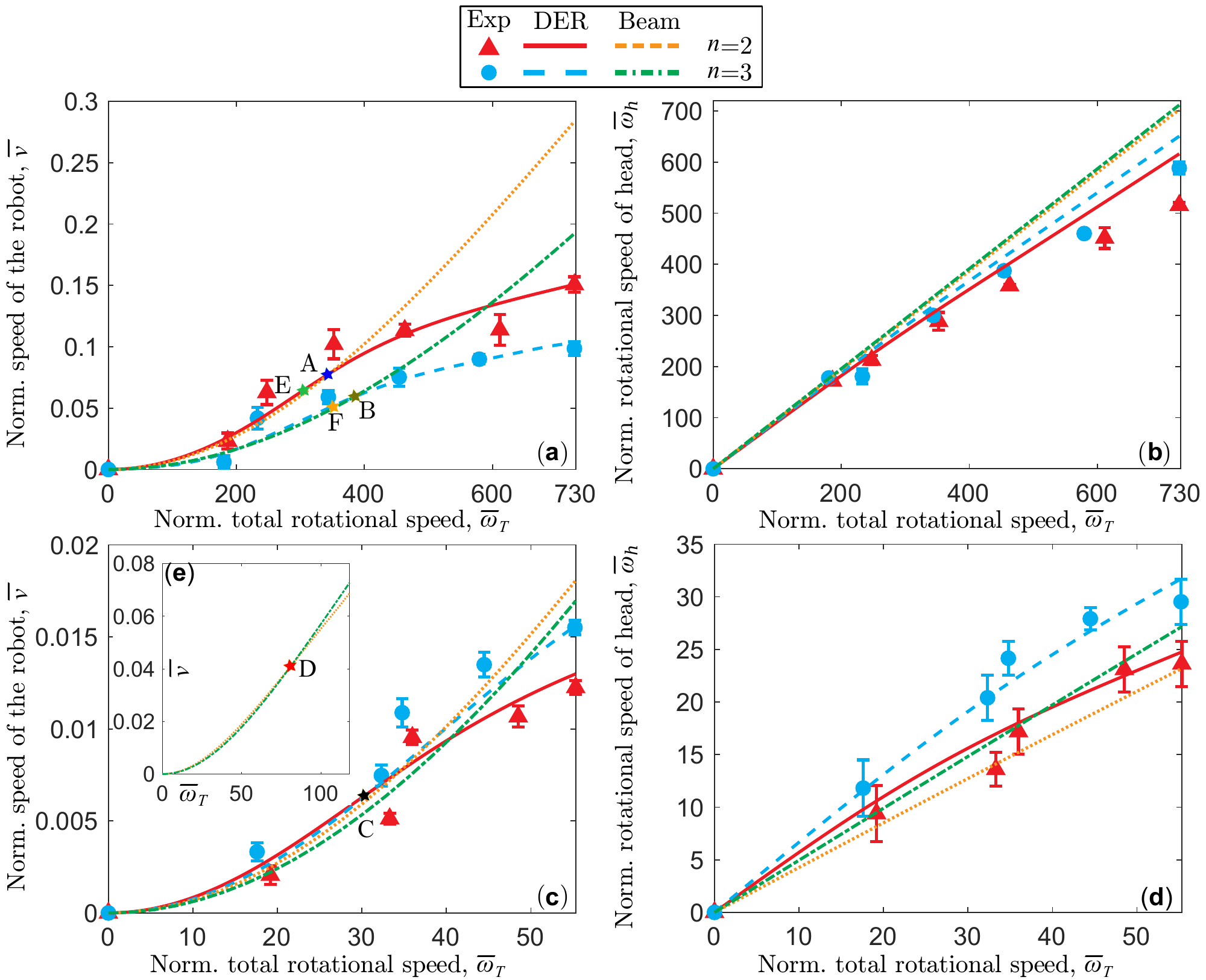}
    \caption{Comparison among the experimental data, DER-based simulation results and nonlinear Euler-Bernoulli beam(\textit{NLB})-based framework prediction. (\textbf{a}) (\textbf{c}) The plot of normalized robot movement speed $\bar{v}$ versus normalized total rotational speed of the robot $\bar{\omega}_T$; \textbf{(a)} Comparison of experimental data, DER-based simulation results and beam-based calculation in design 1, \textbf{(c)} comparison of experimental data and DER-based simulation outcomes of design 2, \textbf{(e)} beam-based framework prediction of the appearance of design 2 after point $D$; (\textbf{b})(\textbf{d}) The plot of normalized rotational speed of the robot head $\bar{\omega}_h$ versus normalized total rotational speed of the robot $\bar{\omega}_T$; (\textbf{b}) Comparison of experimental data, DER-based simulation results and beam-based calculation in design 1, (\textbf{d}) comparison of experimental data, DER-based simulation outcomes and beam-based framework prediction result of design 2.
    % \fix{Separate panel or figure: Zoom into small omega regime and plot beam theory data. Show that beam theory agrees with DER in both the regimes. Then we can say that beam theory captures the two regimes despite its simplicity.}
    }
    \label{fig:SpeedVSFlagellaNumber}
\end{figure*}

The speed of the robot is the rate of change of its position with respect to the corresponding time interval, i.e. $v = \frac{\Delta s}{\Delta t}$. Except in some cases where sometimes the position of the robot remains unchanged (and ``jamming" happens) when the number of flagella is large (e.g. $n = 4$ and $n = 5$), the position of the robot increases proportionally with time. Hence, $v$ is a constant throughout each experiment trial. Since we apply RFT to model the hydrodynamic forces onto the robot from the GM, periodic ``jamming" phenomenon cannot be captured by our simulator. We focus only on robots with $n = 2$ and $n = 3$ flagella that maintain a constant $v$ over time. However, ``jamming" (i.e. the viscosity of the GM in front of the head increases) can be effortlessly modeled as a function of the robot configuration and integrated into the DER-based simulator. 
We exhibit experimental results in Figs.~\ref{fig:SpeedVSFlagellaNumber}(\textbf{a})(\textbf{b}) (design 1) and (\textbf{c})(\textbf{d})(\textbf{e})(design 2). To obtain each data point (($\bar{\omega}_T$, $\bar{v}$) in Fig.~\ref{fig:SpeedVSFlagellaNumber}(\textbf{a})(\textbf{e}) and ($\bar{\omega}_T$, $\bar{\omega}_h$) in Fig.~\ref{fig:SpeedVSFlagellaNumber}(\textbf{b})(\textbf{d}) where $\bar{v} = v \eta_p L^4/(EI)$  and $\bar{\omega}_h = \omega_h \eta_p L^3/(EI) $ ), we randomly selected three separate one-minute sequence from every ten-minute experimental trial, repeated the operation for three different experimental trials, processed the data for every one-minute sequence, and calculated the average and variance (of nine values). Notice that there are no data points between $\bar{\omega}_T = 0 $ and $\bar{\omega}_T \approx 200$ in Figs.~\ref{fig:SpeedVSFlagellaNumber}(\textbf{a})(\textbf{b}) and $\bar{\omega}_T = 0 $ and $\bar{\omega}_T \approx 18$ in Figs.~\ref{fig:SpeedVSFlagellaNumber}(\textbf{c})(\textbf{d}). This is because we observed from experiments that if $\omega_T$ is below a threshold ($\omega^s_T$), the robot would stay stationary ($v = 0$) but started to move continuously otherwise. The threshold changed depending on the robot design, such as the shape, surface friction and inertia of the robot head and flagellar length. For example, $\omega^s_T \approx 50$ rpm ($\bar{\omega}_T \approx 200$) for the robot used in design 1\cite{du2021modeling} while $\omega^s_T \approx 33$ rpm ($\bar{\omega}_T \approx 18$) for the robot in design 2.

Note that the robots for designs 1 and 2 have distinct characteristics, such as the length of flagella $L$, and all data in Fig.~\ref{fig:SpeedVSFlagellaNumber} normalized and hence dimensionless. Normalization involves design parameters, $L$ for example. As a result, though the ranges of $\bar{\omega}_T $ in Fig.~\ref{fig:SpeedVSFlagellaNumber}(\textbf{a})(\textbf{b}) and (\textbf{c})(\textbf{d}) are seemingly different, they really overlap quite a bit ($\omega_T = 0-250$ rpm in Fig.~\ref{fig:SpeedVSFlagellaNumber}(\textbf{a})(\textbf{b}) and $\omega_T = 0-100$ rpm (Fig.~\ref{fig:SpeedVSFlagellaNumber}\textbf{c})(\textbf{d})). To keep the same number of data points, and make $\omega_T$ low enough for Euler-Bernoulli beam theory to apply, we did not include any additional data in Fig.~\ref{fig:SpeedVSFlagellaNumber}(\textbf{c})(\textbf{d}).

\subsection{Parameter fitting for simulations}
\label{sec:paramFitting}
Next, we will show the DER-based and beam-based numerical simulation results for locomotion modeling. Recall that $C_1$ (in Eqs.~\ref{eq:vHead} and~\ref{eq:stokes_drag_refined}) and $C_2$ (in Eqs.~\ref{eq:OHead} and~\ref{eq:stokes_torque_refined}) are two fitting parameters to account for the shape and surface roughness. Additionally, $\mu$, the constant used to quantify the body-granule friction coefficient, is the third fitting parameter. 
As detailed next, $C_1$, $C_2$, and $\mu$ in both designs 1 and 2 were generated from experimental data of 2-flagellar locomotion, and the same values were used in simulations to predict 3-flagellar locomotion. The predicted outcomes were compared to the corresponding experimental data and validated the fitting process.

The normalized total rotational speed of motor, $\bar{\omega}_T$, compares the period of angular rotation to the elasto-viscous relaxation time, $\bar{T} = \eta_p L^4/(EI)$, where $EI$ is the bending modulus of the flagella and $\eta_p$ is the drag coefficient along the perpendicular direction \cite{coq2008rotational}, i.e. $\bar{\omega}_T = {\omega}_T\eta_p L^4/(EI)$. Similarly, the rotational speed of the head is normalized as $\bar{\omega}_h = {\omega}_h\eta_p L^4/(EI)$ and the velocity of the robot is normalized as $\bar{v} = v\eta_p L^3/(EI)$. In Figs.~\ref{fig:SpeedVSFlagellaNumber}(\textbf{a})(\textbf{b}) (for design 1), we plot the normalized speed of the robot, $\bar{v}$, versus the normalized total rotational speed of the motor, $\bar{\omega}_T$ and normalized rotational speed of the head $\bar{\omega}_h$ versus $\bar{\omega_T}$, respectively. All the experimental data, DER-based and beam-based simulation results for $n = 2$ and $n = 3$ are shown in the figures. The experimental data ($\bar{v}$ vs. $\bar{\omega}_T$ and $\bar{\omega}_h$ vs. $\bar{\omega}_T$) for $n = 2$ were adopted to determine the best fit values of the fitting parameters: $C_1 = 2.420, C_2 = 0.039$ and $\mu = 6.828$ (for design 1). These parameter values  were then applied into the numerical simulators to simulate the locomotion performance of $n = 3$. In design 2, the same technique was applied in Figs.~\ref{fig:SpeedVSFlagellaNumber}(\textbf{c})(\textbf{d}) to find the fitting parameters that minimize the fitting error between experimental and DER-based simulation results : $C_1 = 28.750, C_2 = 0.938$ and $\mu = 2.125$ (for design 2). Figs~\ref{fig:SpeedVSFlagellaNumber}(\textbf{a})(\textbf{c}) demonstrate a high degree of agreement between experiments and DER-based simulations. The robot designs for designs 1 and 2 are distinct, and the relevant physical parameters were detailed in Section~\ref{subsec:parameters}.  

The following modeling assumptions can partially cause the slight disparity between the experimental and simulation results:
\begin{itemize}
    \item In this fluid (i.e., the granular medium) structure (i.e., the robot) interaction modeling of robot locomotion, we assume that RFT can characterize the drag from the GM (the fluid).
    \item The structure model (DER) assumes that the flagella are infinitesimally thin elastic rods.
    \item It is assumed that the drag force acting on the head is considered to be linearly proportional to the velocity, and torque on the head is linearly proportional to its angular speed.
\end{itemize}
In addition, inevitably, there are experimental errors, structural defects introduced during fabrication, for instance. Nevertheless, the reasonable consistency between experiments and simulations suggests the validity of RFT in this context. 
%
% \begin{figure*}[t!]
%     \centering
%     \includegraphics[width=0.75\textwidth]{figures/YayunFig6.pdf}
%     \caption{ Results of normalized speed of the robot versus normalized total rotational speed of the motor for robots with four ($n=4$) and five ($n=5$) flagella based on DER simulation and nonlinear beam (\textit{NLB}) framework.
%     }
%     \label{fig:SpeedVSFlagellaLengthandNumber}
% \end{figure*}
\subsection{Speed vs. robot geometry}
\label{subsec:VelocityVSgeometry}
To quantify the effect of geometrical design, e.g. the number $n$ and length $L$ of flagella, we perform a parameter sweep along the angular velocity to systematically study the mechanical response of robots in both designs. 
\subsubsection{Speed vs. number of flagella}
\label{subsubsec:VelocityVSFlagellaNumber}
After performing the first round of experiments and data processing, we observed a counter intuitive phenomenon from Fig.~\ref{fig:SpeedVSFlagellaNumber}(\textbf{a}) that the robot with two flagella ($n = 2$) moved faster than the one with three flagella ($n = 3$), at the same value of $\bar{\omega}_T$. Moreover, the relationship between $\bar{v}$ versus $\bar{\omega}_T$ is nonlinear. We then used the optimal set of fitting parameters described in Section~\ref{sec:paramFitting} to perform the beam-based framework calculation; the results (($\bar{v}$ vs. $\bar{\omega}_T$ and $\bar{\omega}_h$ vs. $\bar{\omega}_T$)) are plotted in Figs.~\ref{fig:SpeedVSFlagellaNumber}(\textbf{a})(\textbf{b}). We denote points $A$ and $B$ in Fig.~\ref{fig:SpeedVSFlagellaNumber}(\textbf{a}) as the intersection of beam-based and DER-based simulation results for $n = 2$ and $n = 3$ in design 1, respectively. As can be observed, the beam-based simulation results are close to DER-based simulation results when $\bar{\omega}_T$ is less than the value at point $A$ ($n = 2$) and point $B$ ($n = 3$), but deviate in other cases. All these observations indicate the significant flagellar deformation and the tight coupling between the head and flagella. 

Nonetheless, the beam-based mechanics analysis framework in Section~\ref{subsubsec:flagellarPropulsion} tells us that if there is no head or the head is fixed along the x-axis (in Fig.~\ref{fig:schematic}), $\textbf{F}_p$ will increase proportionally to the increase in $n$. The propulsive speed of the robot, as illustrated in Fig. 2, is the result of a complex battle between the projection of $\mathbf{v}_t$ along $-x$ (propulsion) and the sum of $\textbf{F}_p$ in Eq.~\ref{eq:stokes_drag_refined} and the projection of $\mathbf{v}_p$ along $-x$ (friction force). Section~\ref{subsubsec:headeffect}, on the other hand, elucidates that the effect induced by the head is one of the reasons for the difference between designs 1 and 2. Notice that $\bar{\omega}_h$ occupies more than $80\%$ of $\bar{\omega}_T$ in Fig.~\ref{fig:SpeedVSFlagellaNumber}($\textbf{b}$), indicating the majority of the increment in $\bar{\omega}_T$ is spent growing $\bar{\omega}_h$. As a result, to achieve design 2, which is predicted by the beam-based framework, we have to slow down the rotational speed of the head. As shown in Fig.~\ref{fig:SpeedVSFlagellaNumber}($\textbf{d}$), the ratio of  $\bar{\omega}_h$ to $\bar{\omega}_T$ in design 2 drops to around $55\%$.  Section~\ref{subsec:robotdesign} details how this was accomplished by updating the robot design in design 2 relative to the one utilized in design 1. We performed experiments with the updated robot design and the results are shown in Figs.~\ref{fig:SpeedVSFlagellaNumber}(\textbf{c})(\textbf{d}). Note that all of the results that pertain to beam theory in Fig.~\ref{fig:SpeedVSFlagellaNumber} are \textit{NLB w/o head}, as illustrated in Section~\ref{subsubsec:headeffect}.
\begin{figure*}[t!]
    \centering
    \includegraphics[width=0.75\textwidth]{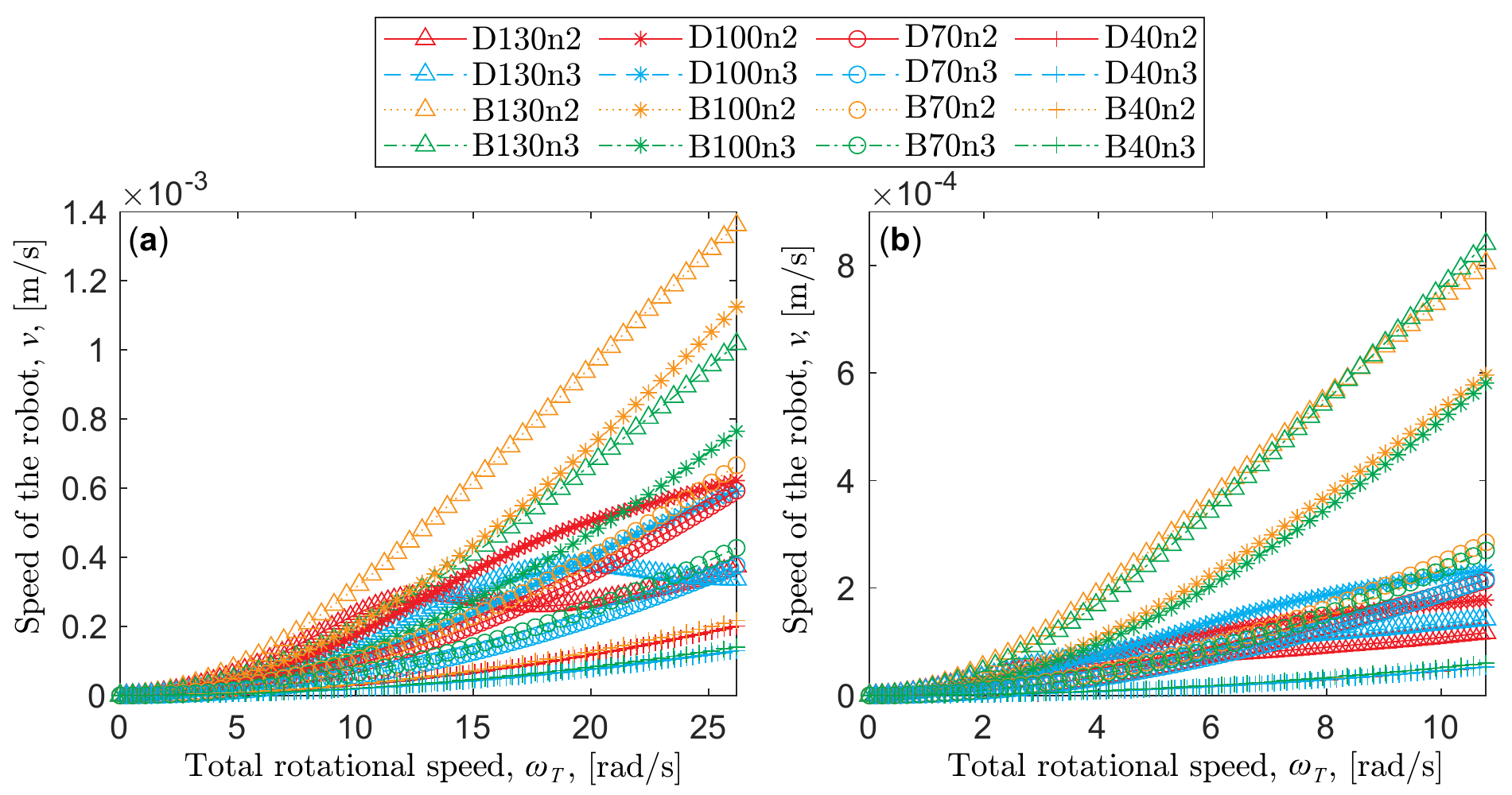}
    \caption{Comparison between DER-based simulation results and Euler-Bernoulli beam-based framework prediction in terms of flagellar number and length. (\textbf{a})(\textbf{b}) The plot of robot movement speed $\bar{v}$ versus total rotational speed of the robot $\bar{\omega}_T$; \textbf{(a)} corresponds to design 1, while (\textbf{b}) corresponds to design 2. In the abbreviated legend, ``D" stands for DER-related results and ``B" represents results from the nonlinear beam theory-based framework; the number following ``D"/``B" represents the length of each flagellum in millimeters(mm), ``n2" denotes two flagella and ``n3" denotes three flagella. For example, ``D130n2" stands for the results of a DER-based simulation of a robot with two 130mm flagella.
    }
    \label{fig:SpeedVSFlagellaLengthandNumber}
\end{figure*}
As mentioned in the last session, when $n$ equals two, the parameter fitting technique was leveraged to match the experiments and simulations. Point $C$ in Fig.~\ref{fig:SpeedVSFlagellaNumber}(\textbf{c}) stands for the conjunction of DER-based simulation results of $n = 2$ and $n = 3$. Experimental data point out that a two-flagellar robot moves slower than a three-flagellar robot while all other parameters remain constant.
There is no evident difference in $\bar{v}$ between $n = 2$ and $n = 3$ simulation results for values of $\bar{\omega}_T$ smaller than the value at point $C$, which was captured by experiments. However, beam-based prediction does not show design 2 in Fig.~\ref{fig:SpeedVSFlagellaNumber}(\textbf{c}) but it demonstrates design 2 when we increase the range of $\bar{\omega}_T$ as shown in Fig.~\ref{fig:SpeedVSFlagellaNumber}(\textbf{e}). Furthermore, our goal of reducing the proportion of $\bar{\omega}_h$ taken from $\bar{\omega}_T$ is verified in Fig.~\ref{fig:SpeedVSFlagellaNumber}(\textbf{d}). In Fig.~\ref{fig:SpeedVSFlagellaNumber}(\textbf{e}), we display the beam-based framework predictions with the fitting parameters $C_1 = 28.750, C_2 = 0.938$ and $\mu = 2.125$ plugged in. Here, $D$ represents the watershed point, beyond which the robot with $n = 3$ moves faster than the robot with $n = 2$ and vice versa. The beam-based prediction follows the same trend with the DER-based simulation in Fig.~\ref{fig:SpeedVSFlagellaNumber}(\textbf{c}) even though the watershed points are drastically different. Also, the magnitudes of $\bar{v}$ and $\bar\omega_h$ predicted by the beam-based framework are a lot larger than those obtained from experiments and DER-based simulation. This further supports our assertion that robot locomotion involves large flagellar deformation and close interaction between the head and flagella. In summary, the simple beam-based analytical framework qualitatively captures the relationship between the speed of the robot and the number of flagella, though not quantitatively.

One more thing that needs our attention is the nonlinear increment of slope in Fig.~\ref{fig:SpeedVSFlagellaNumber}(\textbf{a}). Both slopes for $n = 2$ and $n = 3$ initially increase in magnitude until they reach a point after which they begin to decrease in magnitude. We refer to the point for $n = 2$ as point $E$ and the one for $n = 3$ as point $F$. The points $E$ and $F$ are conspicuous because, at these points, the unit magnitude increase in $\bar{\omega}_T$ turns into the steepest increase in $\bar{v}$, indicating the maximum efficiency of the robot locomotion is reached. The same phenomenon happens in design 2 (see Fig.~\ref{fig:SpeedVSFlagellaNumber}(\textbf{c})). Last but not least, point $E$ is near point $A$ and $F$ is in the neighborhood of point $B$. This could mean that an increase in large flagellar deformation implies a decline in locomotion efficiency. 
\subsubsection{Speed vs. flagellar length}
\label{subsubsec:VelocityVSFlagellaLength}
 As previously stated, the data in Fig. ~\ref{fig:SpeedVSFlagellaNumber} has been normalized. To help readers visualize the physical scenarios, we plot the simulation results from DER and \textit{NLB} in Fig.~\ref{fig:SpeedVSFlagellaLengthandNumber}, which depicts the relationship between the speed of the robot $v$ in terms of each flagellar length, $L$, and the total rotational speed of the motor, $\omega_T$. As illustrated in Fig.~\ref{fig:SpeedVSFlagellaLengthandNumber}(a)(design 1), DER demonstrates that when the number of flagella, i.e. $n$ is fixed, the distinction between DER and nonlinear beam theory (\textit{NLB}) becomes more discernible as flagella become longer. When $\omega_T$ is small, DER results indicate that $v$ grows as the value of $L$ increases, but this trend breaks down when $\omega_T$ is big. In comparison, \textit{NLB} behaves more consistently and linearly, i.e. when $n$ is constant, $v$ increases with the growth in $L$. These observations make sense in light of the fact that we previously acknowledged that DER is capable of capturing the nonlinearity of flagellar deformation, whereas \textit{NLB} is not. The DER results displayed in Fig.~\ref{fig:SpeedVSFlagellaLengthandNumber}(b)(design 2) show that as $L$ increases, the difference between $v$ of $n=3$ and $n=2$ does not always increase, but in the range of $\omega_T$ presented, $v$ of $n=3$ is greater than $v$ of $n=2$. In comparison to Fig. ~\ref{fig:SpeedVSFlagellaNumber}(c), \textit{NLB} predicts the appearance of design 2 after a threshold is reached. However, this threshold becomes smaller when $L$ is larger. In reality, we tested designs 1 and 2 on robots with $n=4$ and $n=5$ flagella. Nonetheless, ``jamming" happened from time to time randomly. As a result, only simulation data are presented here.
%
% \subsection{}
% \label{subsec:}
\subsection{Deflection of beam end}
\label{subsec:deflection}
\begin{figure}
    \centering
    \includegraphics[width=0.75\columnwidth]{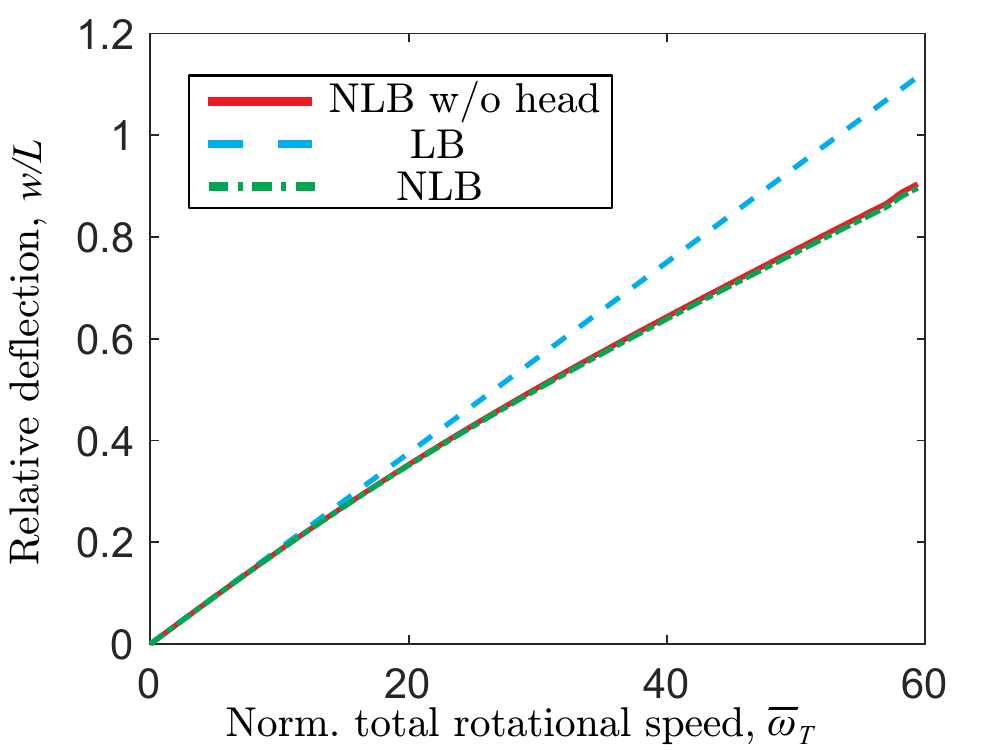}
    \caption{Comparison of the relative deflection of flagella  $w/L$ versus the normalized total rotational speed of the motor, $\bar{\omega}_T$, in regimes \textit{NLB w/o head}, \textit{LB}, and \textit{NLB}. Physical parameters are identical to those in Section~\ref{subsec:parameters} for design 1  and to the fitting parameters in Section~\ref{sec:paramFitting} for design 1.
    }
    \label{fig:comparison}
\end{figure}

In the last session, we state that the simulation results from the beam-based framework (\textit{NLB w/o head} design in Section~\ref{subsubsec:headeffect}) are qualitatively compatible with both experimental and DER-based simulation results. We also emphasize the importance of substantial flagellar deformation in relation to the performance of robot locomotion. Consequently, in Fig.~\ref{fig:comparison}, we compare beam-based deflection in three regimes, \textit{LB}, \textit{NLB}, and \textit{NLB w/o head} as described in Section~\ref{subsubsec:headeffect}.
The simulation parameters are identical to those for design 1 in Section~\ref{subsec:parameters} and to the fitting parameters for design 1 in Section~\ref{sec:paramFitting}.
As seen in Eq.~\ref{eq:deflection}, the effect of the head in the flagellar deflection is ignored in regime \textit{LB}, so the rotational speed of the flagella, $\omega_t$ is chosen as the independent variable. To see more noticeable difference in flagellar deflection among three regimes, we picked the scenario with larger flagellar deformation, design 1, and the corresponding fitting parameters to compute $\omega_h$ in \textit{NLB} and \textit{NLB w/o head} regimes. As seen in Fig.~\ref{fig:comparison}, 
% \omega_t = 0 - 17$ rpm and as indicated in Fig.~\ref{fig:SpeedVSFlagellaNumber} (\textbf{b}), this range corresponds to $\omega_T = 0-180$ rpm. The range of $\omega_T$ encompasses most of the experiments. 
%
when $\omega_t$ is relatively small ($\omega_t \lesssim 5$ rpm and $\bar{\omega}_T \lesssim 20$), the flagellar deformation drops to the \textit{LB} regime introduced in Section~\ref{subsubsec:headeffect} and the difference in deflection of the end of the beam (flagella) among the three regimes is subtle ($\lesssim 5\%$). Moreover, as can be observed in Fig.~\ref{fig:comparison}, the deflection of the end of the beam in regime \textit{NLB w/o head} is almost always equal to that in \textit{NLB} throughout the range of $\bar{\omega}_T$. This corroborates our statement that the head velocity-induced effect could be ignored in Section~\ref{subsubsec:headeffect}.

% The normalized terms are as follows:
% (1) normalized rotational speed 
% \begin{equation}
%     \Bar{\omega_{[]}} = \frac{\omega_{[]}\mu L^4}{EI}
% \end{equation}
% (2) normalized propulsive force
% \begin{equation}
%     \Bar{F_p} = \frac{F_p L^2}{EI}
% \end{equation}

% \subsection{Body Rotation Rate}

% \subsection{Efficiency}

\section{Conclusion}
\label{sec:conclusion}
In this study, we designed a low-cost experimental setup, an untethered robot actuated by multiple soft flagella moves in granules. Meanwhile, we developed two numerical simulators, one based on Euler-Bernoulli beams and the other on discrete differential geometry (DDG) framework, to simulate the performance of the articulated locomotion in granular media (GM). Both numerical tools use resistive force theory (RFT) to model the drag force exerted by the GM on the robot and Stokes\rq{} law to model the external force/moment applied to the robot head by the GM. Initially, experiments unveiled a counterintuitive phenomenon: the robot's speed decreases as the number of flagella increases (design 1). However, the beam-based simulator predicts the existence of the converse case (design 2) on the condition that the rotational speed of the head is suppressed to be zero. The robot design was then modified to slow down the rotational speed of the robot head. 

The DDG-based simulator models the robot into a composition of Kirchhoff elastic rod and discretizes it into a series of mass-spring systems. As a result, the elastic energy of the robot structure is the linear sum of the discrete elastic energies associated with each spring, which include the stretching and coupled bending and twisting energies. At each time step, the equations of motion formulated can be summarized as follows: at each degree of freedom (DOF), the sum of elastic forces (i.e., the negative gradient of elastic energies) and external forces equals the lumped mass multiplied by the acceleration of that DOF. Especially, the actuation of the robot (i.e., the rotational speed of the motor) is modeled by a time-varying natural strain (twist) at the node standing for the head. This method enables us to simulate the robot locomotion fully implicitly. 

Both beam-based and DDG-based frameworks capture designs 1 and 2 successfully. Although the beam-based simulator can quantify only the trend in both designs while the DDG-based method accurately reproduces experiments, it is still exhilarating because the simple beam theory-based analysis depicts the complicated locomotion system. Simultaneously, the discrepancy between beam-based prediction and experiments indicates the large flagellar deformation and nontrivial coupling between the head and flagella. Additionally, both simulators, validated experimentally, shed light on the highly nonlinear functional relationship between the locomotion performance, such as speed and efficiency, and its physical parameters, such as the number of flagella. The nonlinear dependency of the speed of the robot concerning the rotational speed of the motor necessitates the development of a design tool for optimal control of this class of robots. 
Last but not least, thanks to the simplicity of the beam-based analysis framework and the computational efficiency of the DDG-based simulator, they can be exploited to perform parametric studies and identify the optimal design and control this class of articulated robots as long as they locomote through the GM. We chose a smooth and soft GM with a large particle diameter. Scaling power requirements for locomotions moving in harsher GM-like sand will be an interesting challenge. 
%%%%%%%%%%%%%%%%%%%%%%%%%%%%%%%%%%%%%%%%%%%%%%%%%%%%%%%%%%%%%%%%%%%%%%%%%%
\section*{Acknowledgements} 
This work was supported by the Henry Samueli School of Engineering and Applied Science, University of California and the National Science Foundation (Award \# IIS-1925360).

\section*{Author Disclosure Statement}
No competing financial interests exist.

\section*{Supplementary Material}
Supplementary Video S1

% \bibliography{apssamp}% Produces the bibliography via BibTeX.

\end{document}